\documentclass[conference]{IEEEtran} 

\usepackage{times}

\usepackage[numbers,sort&compress]{natbib}
\usepackage{multicol}
\usepackage[bookmarks=true]{hyperref}

\title{
HyP-DESPOT: A Hybrid Parallel Algorithm for\\ Online Planning under Uncertainty
}
\author{Panpan Cai, Yuanfu Luo, David Hsu and Wee Sun Lee}

\usepackage{pdfsync}
\usepackage[subrefformat=parens,labelformat=parens]{subfig}
\usepackage{graphicx}
  \graphicspath{{Figures/}}
  \DeclareGraphicsExtensions{.pdf,.jpg,.png,.eps}
\usepackage{pdfpages}
\usepackage{bm}
\usepackage{amsmath,amssymb}
\usepackage{algorithm}
\usepackage{algorithmicx}
\usepackage{algpseudocode}
\usepackage{xspace}
\usepackage{booktabs}
\usepackage{amsthm}

\newcommand{\secref}[1]{Section~\ref{#1}}
\renewcommand{\eqref}[1]{(\ref{#1})}
\newcommand{\figref}[1]{Fig.~\ref{#1}}

\newcommand{\tabref}[1]{Table~\ref{#1}}

\newcommand{\eg}{\textrm{e.g.}}
\newcommand{\etc}{\textrm{etc.}}

\DeclareMathOperator*{\argmax}{arg\,max} 
\newcommand{\mean}{\ensuremath{\mathbb{E}}\xspace}
\newcommand{\pr}{\ensuremath{p}}

\newlength{\citeskipup}
\newlength{\citeskipdown}

\usepackage{color}
\definecolor{fullred}{rgb}{0.95,.0,.1} 
\newcounter{cmt}

\newcommand{\nscen}{\ensuremath{K}\xspace}
\newcommand{\nscensymbol}{\ensuremath{K}}
\newcommand{\scenario}{\ensuremath{{\boldsymbol{\phi}}}\xspace}

\newcommand{\scenariosetnode}[1]{\ensuremath{{\boldsymbol{\Phi}_{#1}}}\xspace}
\newcommand{\scenariosetsize}[1]{\ensuremath{|\scenariosetnode{#1}|}\xspace}

\newcommand{\randnum}{\ensuremath{\varphi}\xspace}

\newcommand{\state}{\ensuremath{s}\xspace}
\newcommand{\scurr}{\ensuremath{s}\xspace}
\newcommand{\snext}{\ensuremath{s'}\xspace}
\newcommand{\obs}{\ensuremath{z}\xspace}
\newcommand{\act}{\ensuremath{a}\xspace}

\newcommand{\optact}{\ensuremath{a^*}\xspace}
\newcommand{\optobs}{\ensuremath{z^*}\xspace}
\newcommand{\obsprob}[3]{\ensuremath{O({#1},{#2},{#3})}\xspace}
\newcommand{\obsfun}{\ensuremath{O}\xspace}
\newcommand{\transfun}[3]{\ensuremath{T({#1},{#2},{#3})}\xspace}

\newcommand{\rfun}[2]{\ensuremath{R({#1},{#2})}\xspace}
\newcommand{\rfunsymbol}{\ensuremath{R}\xspace}

\newcommand{\stateset}{\ensuremath{S}\xspace}
\newcommand{\obsset}{\ensuremath{Z}\xspace}
\newcommand{\actset}{\ensuremath{A}\xspace}
\newcommand{\sinit}{\ensuremath{s_0}\xspace}

\newcommand{\belief}{\ensuremath{b}\xspace}
\newcommand{\node}{\ensuremath{b}\xspace}
\newcommand{\newbelief}{\ensuremath{b'}\xspace}
\newcommand{\newnode}{\ensuremath{b'}\xspace}
\newcommand{\leafnode}{\ensuremath{b}\xspace}
\newcommand{\rootnode}{\ensuremath{b_0}\xspace}
\newcommand{\pol}{\ensuremath{\pi}\xspace}

\newcommand{\defaultpol}{\ensuremath{\pi_0}\xspace}

\newcommand{\polvalue}{\ensuremath{V}\xspace}

\newcommand{\ascale}{\ensuremath{c_a}\xspace}
\newcommand{\oscale}{\ensuremath{c_o}\xspace}

\newcommand{\timestep}{\ensuremath{t}\xspace}
\newcommand{\depth}{\ensuremath{\Delta}\xspace}
\newcommand{\maxsearchdepth}{\ensuremath{D}\xspace}
\newcommand{\nvisit}[1]{\ensuremath{N({#1})}\xspace}
\newcommand{\nvisita}[2]{\ensuremath{N({#1},{#2})}\xspace}
\newcommand{\ubsymbol}{\ensuremath{u}\xspace}
\newcommand{\ub}[1]{\ensuremath{\ubsymbol({#1})}\xspace}
\newcommand{\uba}[2]{\ensuremath{\ubsymbol({#1},{#2})}\xspace}
\newcommand{\iniub}[1]{\ensuremath{\ubsymbol_0({#1})}\xspace}
\newcommand{\lbsymbol}{\ensuremath{l}\xspace}
\newcommand{\lb}[1]{\ensuremath{\lbsymbol({#1})}\xspace}
\newcommand{\inilb}[1]{\ensuremath{\lbsymbol_0({#1})}\xspace}
\newcommand{\gap}[1]{\ensuremath{\epsilon({#1})}\xspace}

\newcommand{\augub}[2]{\ensuremath{u^{+}({#1},{#2})}\xspace}
\newcommand{\augweu}[1]{\ensuremath{E^{+}({#1})}\xspace}
\newcommand{\weu}[1]{\ensuremath{E({#1})}\xspace}
\newcommand{\vloss}{\ensuremath{\zeta}\xspace}
\newcommand{\dsmodelf}[3]{\ensuremath{g({#1},{#2},{#3})}\xspace}
\newcommand{\dsmodel}{\ensuremath{g}\xspace}
\newcommand{\beliefupdate}{\ensuremath{\tau}\xspace}

\newcommand{\accel}{{\small \textsc{Accelerate}}}
\newcommand{\decel}{{\small \textsc{Decelerate}}}
\newcommand{\maintain}{{\small \textsc{Maintain}}}
\newcommand{\sample}{{\small \textsc{Sample}}\xspace}
\newcommand{\sense}{{\small \textsc{Sense}}\xspace}
\newcommand{\good}{{\small \textsc{Good}}\xspace}
\newcommand{\bad}{{\small \textsc{Bad}}\xspace}

\newcommand{\occupied}{{\small \textsc{Occupied}}\xspace}
\newcommand{\free}{{\small \textsc{Free}}\xspace}

\newcommand{\element}{within-step\xspace}
\newcommand{\elementlevel}{within-step\xspace}
\newcommand{\Elementlevel}{Within-step\xspace}
\newcommand{\RockSample}{Rock Sample\xspace}
\newcommand{\RockSampleTight}{Rock Sample}

\IEEEoverridecommandlockouts                              

\begin{document}

\maketitle

\begin{abstract}


  Planning under uncertainty is critical for robust robot performance in
  uncertain, dynamic environments, but it incurs high computational
  cost. State-of-the-art online search algorithms, such as DESPOT, have vastly
  improved the computational efficiency of planning under uncertainty and made
  it a valuable tool for robotics in practice.  This
  work takes one step further by leveraging both CPU and GPU parallelization
  in order to achieve near real-time online planning performance for complex
  tasks with large state, action, and observation spaces.  Specifically, we propose Hybrid Parallel DESPOT
  (HyP-DESPOT), a massively parallel online planning algorithm that integrates
  CPU and GPU parallelism in a multi-level scheme.  It performs parallel
  DESPOT tree search by simultaneously traversing multiple independent paths
  using multi-core CPUs and performs parallel Monte-Carlo simulations at the
  leaf nodes of the search tree
  using GPUs.  Experimental results show that HyP-DESPOT speeds up online
  planning by up to several hundred times, compared with the original DESPOT
  algorithm, in
  several challenging robotic tasks in simulation.

\end{abstract}
\IEEEpeerreviewmaketitle

\section{Introduction}

\begin{figure}
	\includegraphics[width=0.5\textwidth]{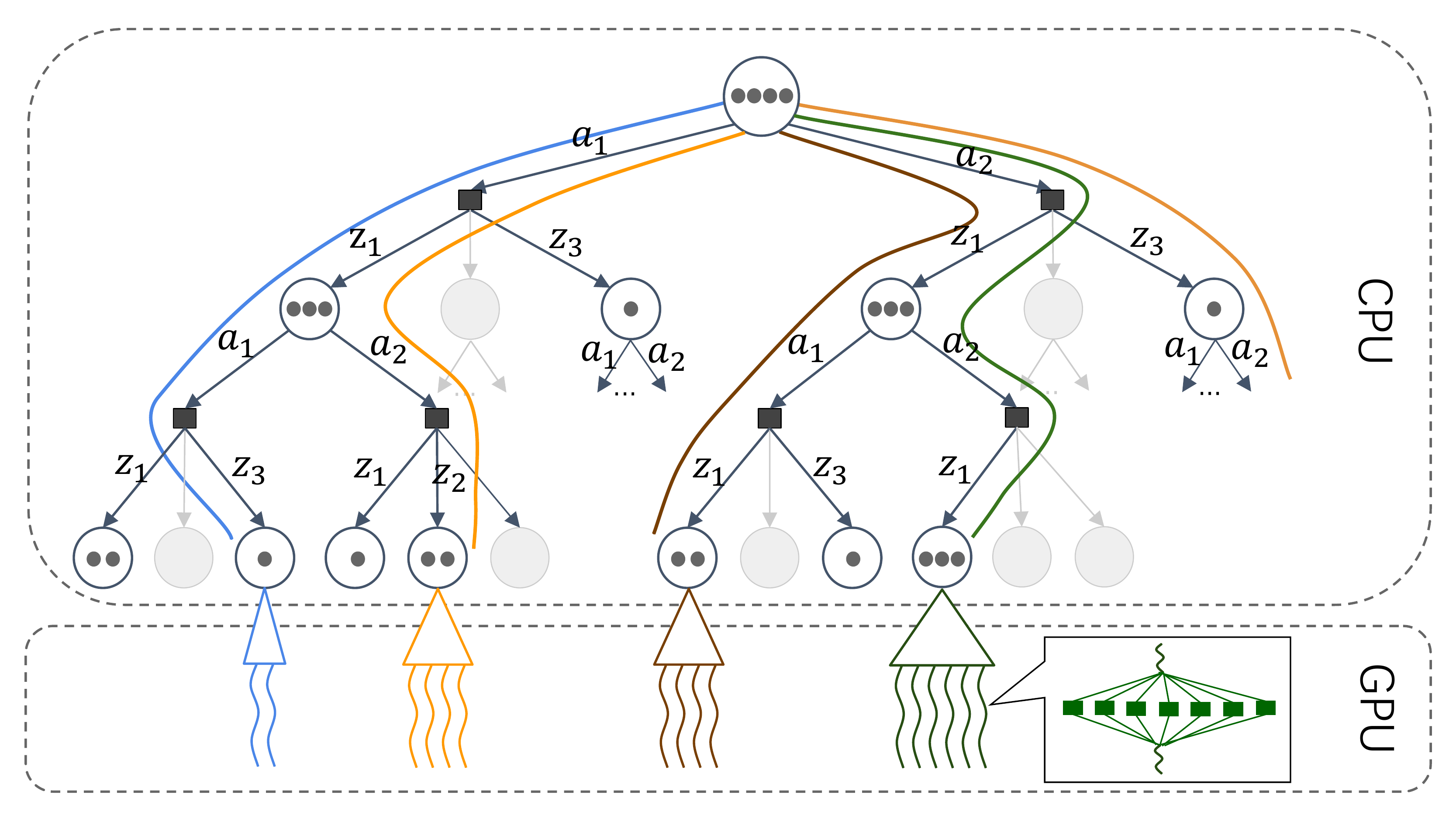}
	\caption{An overview of HyP-DESPOT.
         Each node of the belief
          tree (gray) represents a belief. A parent node and a child node, with
          associated beliefs $b$ and $b'$ respectively, 
          are connected by an action-observation pair $(a,z)$, indicating that
          the belief transitions from $b$ to $b'$, when a robot, with initial
          belief $b$, takes actions $a$ and receives observation $z$.
          The DESPOT tree (black) is a  sparse subtree of the belief tree
          and contains only branches reachable under a set of sampled
          scenarios (black dots). HyP-DESPOT integrates CPU and GPU
          parallelism:
          multi-threaded parallel tree search
          (colored paths) in the
          CPUs, massively parallel Monte Carlo simulation at the leaf nodes in
          the GPUs, and fine-grained GPU parallelization
          within a simulation step (inset figure).  
        }
	\label{fig:Overview}
\end{figure}

As robots move towards uncontrolled natural human environments in our daily
life---at home, at work, or on the road---they face a plethora of uncertainties:
imperfect robot control, noisy sensors, and fast-changing environments. A key
difficulty here is \emph{partial observability}: the system 
states are not known
exactly.  A principled way of handling partial observability  is to capture the
uncertainties in a \emph{belief}, which is a probability distribution over
states, and reason about the effects of robot actions, sensor
information, environment changes on the belief.  To formalize this, a planning
algorithm performs look-ahead search in a \emph{belief tree}, in which each
tree node represents a belief, and parent and child nodes are connected by 
action-observation pairs (\figref{fig:Overview}).  While the belief tree search
is conceptually simple, it is computationally intractable in the worst case,
as the number of states or the planning time horizon increases.

DESPOT~\cite{DESPOT} is a state-of-the-art belief tree search algorithm for
on planning under uncertainty. To overcome the computational challenge, DESPOT
samples a set of ``scenarios'' and constructs incrementally---via heuristic tree
search and Monte Carlo simulation---a \emph{sparse} belief tree, which contains
only branches reachable under the sampled scenarios (\figref{fig:Overview}).  The sparse tree is
provably near-optimal~\cite{DESPOT}, and DESPOT has shown strong
performance in various robotic tasks, including autonomous
driving~\cite{Bai_2015} and manipulation~\cite{Li_2016}.


Our goal here is to scale up DESPOT further through parallelization and achieve
near real-time performance for online planning under uncertainty in complex
tasks with large state, action, and observation spaces. 
Specifically, we propose \emph{Hybrid Parallel DESPOT} (HyP-DESPOT), which
exploits both multi-core CPUs and GPUs to form a multi-level
parallelization scheme for DESPOT. 

First,
HyP-DESPOT uses multiple CPU threads to perform parallel tree search by
simultaneously traversing many paths.
The CPU threads provide the flexibility to handle the irregularity of tree
search for parallelization.
The key issue here is to distribute the
threads over a diverse set of tree paths, with minimum communication
among the threads.

Second, HyP-DESPOT uses GPUs to perform massively parallel Monte Carlo
simulations at the belief tree node level, the action level, and the scenario
level.  Further, a complex system often consists of multiple components, \eg,
multiple robots or humans in an interactive or collaborative
setting. HyP-DESPOT factors the dynamics model and the observation model of
such a system in order to extract additional opportunities for GPU
parallelization at a fine-grained level. Since the simulations are
independent, parallelization is conceptually straightforward.  However, GPUs
suffer from high memory access latency and low single-thread arithmetic
performance.  Parallel simulation and parallel tree search must be integrated
in order to generate sufficient parallel workload and benefit from large-scale
GPU parallelization.



To the best of our knowledge, HyP-DESPOT is the first massively parallel
algorithm for online planning under uncertainty.  Our experiments show
that HyP-DESPOT achieves significant speedup and
better  solutions,  compared with the original  DESPOT algorithm.

\section{Background}

\subsection{Online Planning under Uncertainty}
A robot operates in a partially observable stochastic environment.
The robot has state space \stateset, action space \actset, and observation
space \obsset.
We model the robot's stochastic dynamics with a probability  function
$T(s, a, s') = \pr(s' | s, a)$ for $s,s'\in\actset$ and $a\in\actset$. We
model the noisy sensors with another probability  function
$\obsfun(s', a, z) = \pr(z | a, s')$, for $s'\in \stateset$, $a\in \actset$,
and $z\in \obsset$. 

There are two distinct approaches to planning under uncertainty: offline (\eg,
\cite{PBVI,HSVI,SARSOP,MCVI}) and
online (\eg,\cite{AEMS,POMCP,DESPOT}).
Offline planning leverages offline computation to reason about all
future contingencies in advance and achieves faster execution time
online. In contrast, online planning focuses the computation
on the contingency currently encountered and scales up to much more complex
tasks. 

For online planning under uncertainty, a robot computes an action at each time
step and interleaves planning and action execution. To determine the best action
at the current belief~\belief, we
perform lookahead search in a belief tree rooted at \belief~(\figref{fig:Overview}).  The search
optimizes the value over all policies:
\begin{equation}\label{eq:1}
 \pol^*(\belief)=\argmax\limits_{\pol}\polvalue_{\pol}(\belief).
\end{equation}
A  \emph{policy} \pol specifies the robot action at every belief, and 
the \emph{value} of \pol at a belief $b$,  $\polvalue_\pol(b)$, is the expected total
discounted reward of executing the policy, with initial
belief $b$:
\begin{equation}
\polvalue_{\pol}(\belief)=\mean\Biggl(\sum\limits_{\timestep=0}^{\infty}
  \gamma^\timestep \rfun{\state_\timestep}{\pol(\belief_\timestep)} \biggr| \belief_0=\belief \Biggl),
\end{equation}
where
$\rfunsymbol(s,a)$ is a real-valued reward function designed to capture
desirable robot behaviors 
 and $\gamma$ is a discount factor expressing  the
preference for immediate rewards over future ones. 
The robot then executes the action $\act=\pol^*(b)$
and  receives an observation~\obs. We update the belief
by incorporating the information in $a$ and $z$, according to  the Bayes' rule:
\begin{equation}
\newbelief(\snext)=\beliefupdate(\belief,\act,\obs)=\eta
\obsprob{\snext}{\act}{\obs}\sum\limits_{\scurr \in \stateset}
\transfun{\scurr}{\act}{\snext}\belief(\scurr), 
\end{equation}
where $\eta$ is a normalization constant.  The new belief \newbelief then
becomes the entry point of the planning cycle for the next time step.

While  belief tree search incurs high computational cost, Monte Carlo sampling
is a powerful idea to make it efficient in practice. 
Early examples include the roll-out algorithm~\cite{Rollout}, sparse
sampling~\cite{SparseSampling}, hindsight optimization~\cite{Hindsight}, and
AEMS~\cite{AEMS}.  Two state-of-the-art online planning algorithms,
POMCP~\cite{POMCP} and DESPOT~\cite{DESPOT}, both make use of Monte Carlo
sampling. 
POMCP performs Monte Carlo tree search (MCTS) on the belief tree and
uses the partially observable UCT algorithm (PO-UCT) to trade off exploration
and exploitation.
DESPOT performs anytime heuristic search in a sparse belief tree conditioned
on a set of  sampled scenarios.
Both POMCP and DESPOT solve moderately large planning tasks under uncertainty
efficiently, while DESPOT achieves significantly better worst-case
performance. More importantly, DESPOT offers better opportunities for
parallelization, as it generates a large number of Monte Carlo simulations,
each corresponding to a sampled scenario, and process them 
simultaneously rather sequentially, as POMCP does.

\subsection{Parallel Planning under Uncertainty}
Planning under uncertainty can be formalized as a \emph{Markov decision process}
(MDP)
if the system state is fully observable, or as a \emph{partially observable
  Markov decision process} (POMDP)
if the system state is not fully observable~\cite{RusselNorvig}.
Parallelization is a powerful tool that has been exploited
to speed up both MDP planning
\cite{Chaslot_2008,Rocki_2011,Barriga_2014,Johnson_2016} and
offline POMDP planning \cite{GPOMDP, p_PBVI}. 

The main focus of parallel MDP planning is parallel MCTS:
leaf parallelization \cite{RootLeafP}, root parallelization \cite{RootLeafP}, and tree
parallelization~\cite{Chaslot_2008}. 
Leaf parallelization performs multiple roll-outs from leaf tree nodes in
parallel. Root parallelization builds multiple trees in
parallel to select the best action. Both use multiple CPU threads.
One may also combine leaf and root parallelism and exploit large-scale
 GPU parallelization. 
\citet{Rocki_2011} proposes a block
parallelism scheme, which uses GPUs to parallelize roll-out requests from
multiple trees. \citet{Barriga_2014} extends the idea to a multi-block
parallelism scheme by additionally expanding the children of leaf nodes.
However, increasing the number of trees in root
parallelization or the number of roll-outs
in leaf parallelization often has limited benefits,
because the multiple tree searches are
independent without information sharing and the same computation is repeated
many times.
Tree parallelization addresses this issue by cooperatively searching a shared
tree using multiple CPU threads. The challenge here is to minimize the 
 communication overheads.
 HyP-DESPOT exploits both tree parallelization and leaf parallelization, and
 integrates them in a CPU-GPU hybrid parallel model for belief tree search.


 Offline POMDP planning computes beforehand a policy for all contingencies,
 thus inducing a huge number of independent tasks for parallelization.  gPOMDP
 \cite{GPOMDP} parallelizes the Monte Carlo value iteration (MCVI)
 algorithm~\cite{MCVI} by performing Monte Carlo simulations for multiple
 beliefs, candidate actions, policy graph nodes, \etc, in parallel in GPUs.  A
 similar idea \cite{p_PBVI} is used to parallelize the point-based value
 iteration (PBVI) algorithm~\cite{PBVI}.

 Offline planning has almost unlimited offline computation time to derive a
 solution. In contrast, online robot planning under uncertainty is usually
 given a small fixed amount of time to choose the best action in real time.
 Parallelism is much more important for online planning, in order to scale up
 to complex tasks, but is rarely explored. Our work aims to fill this gap.

\section{Hybrid Parallel DESPOT} 

\subsection{DESPOT}

For completeness, we provide a brief summary of the DESPOT algorithm.
See~\cite{DESPOT} for details.
To overcome the computational challenge of
online planning under uncertainty,  
DESPOT samples a small finite set of \nscen scenarios as representatives of
the future.
Each scenario, $\scenario=(\sinit, \randnum_1,\randnum_2,...)$, contains a
sampled initial state \sinit and random numbers $\randnum_1,\randnum_2,...$,
which determinize the uncertain outcomes of future actions and observations. 

A DESPOT tree is a sparse belief tree conditioned  on the sampled scenarios
(\figref{fig:Overview}). Each node of the tree contains a set of scenarios,
whose starting states form an approximate representation of a belief.
The tree starts with from a initial belief. It branches on all actions, but
only on observations encountered under the sampled scenarios.  

The DESPOT algorithm performs anytime heuristic search and 
constructs the tree incrementally by iterating on the three key steps below.

\subsubsection{Forward Search}\label{Forward_search}
DESPOT starts from the root node \rootnode and searches a single path down to
expand the tree. At each node along the path, DESPOT chooses an action branch
and an observation branch optimistically according to the heuristics defined
by an upper bound and a lower bound value, \ubsymbol and \lbsymbol.
\subsubsection{Leaf Node Initialization}
Upon reaching a leaf node \node, DESPOT fully expands it for one level using
all actions and the observations encountered under the scenarios visiting
\node. It then initializes the upper and lower bounds for the new nodes, by
performing a large number of Monte Carlo simulations.
\subsubsection{Backup}
After creating the new nodes, the algorithm traverses the path all the way
back to the root and updates  the upper and
lower bounds for all nodes along the path, according to the Bellman's principle:
\begin{eqnarray}\label{eqn:backup}
\fontsize{9}{10}
\polvalue(\node)=\max_{\act \in A}\left\{\frac{1}{\scenariosetsize{\node}}
	\sum_{\scenario \in \scenariosetnode{\node}} \rfun{\scurr_{\scenario}}{\act}+\gamma
	\sum_{\obs \in \obsset_{\node,\act}}\frac{\scenariosetsize{\newnode}}{\scenariosetsize{\node}}\polvalue(\newnode)\right\}
\end{eqnarray}
where \polvalue stands for both the upper bound and the lower bound value, and $\newnode=\beliefupdate(\node, \act, \obs)$ represents a child node of $\node$. 


DESPOT repeats the three steps until the gap between the upper and
lower bounds at \rootnode is sufficiently small or the maximum time limit is reached. 

\subsection{HyP-DESPOT Overview}
We want to parallelize all key steps of DESPOT, but they exhibit 
different structural properties for parallelization. The two tree search
steps, forward search and back-up, are irregular; leaf node
initialization, which consists of many identical Monte Carlo simulations with
different initial states, is
regular and embarrassingly parallel.
HyP-DESPOT builds a CPU-GPU hybrid parallel model to treat them separately.
It uses the more flexible CPU threads to handle the two irregular tree search
steps. It uses massively parallel GPU threads to handle the embarrassingly
parallel Monte Carlo
simulations for leaf node initialization.   


GPUs, however, suffers from high memory access latency and low single-thread
arithmetic performance, compared with CPUs.
The main memory latency for GPUs is usually
400$\sim$800 clock cycles \cite{CUDAGM}, while it can be around 15 clock cycles for CPUs \cite{Intel}.
Double-precision arithmetic
instructions on GPUs are also several times lower than those on CPUs
\cite{CUDA,Intel}. Efficient GPU parallelization requires massively parallel
tasks to fully utilize the GPU threads and amortize the latency penalties.

HyP-DESPOT integrates CPU-based parallel tree search and GPU-based parallel
Monte Carlo simulations in a multi-level scheme (\figref{fig:Overview}).
Specifically, HyP-DESPOT launches multiple CPU threads to
simultaneously search different paths and discover leaf nodes.  At the same
time, It relies on the GPU threads to takes over these leaf nodes, expand
them,  and initialize their children through massively parallel Monte Carlo
simulations.
Further, HyP-DESPOT factors the dynamics model and the observation model
within a single simulation step and simulates the factored elements in
parallel, in order to maximally exploit GPU parallelization.
The next two subsections present details on 
the  parallel tree search  (\secref{tree_search}) and parallel Monte Carlo
simulations  (\secref{leaf_expansion_initialization}).

\subsection{Parallel DESPOT Tree Search} \label{tree_search}
The key of parallel DESPOT tree search is to effectively distribute CPU threads across the tree. HyP-DESPOT applies exploration bonuses on the original heuristics in DESPOT to achieve this. In particular, HyP-DESPOT uses a modified PO-UCT algorithm to select an action branch and uses a virtual loss mechanism to select an observation branch for a specific CPU thread.

\subsubsection{Heuristics in DESPOT} \label{Heuristics}
For the completeness of this paper, we first describe the original heuristics used in DESPOT.
At each node \node, DESPOT always traverse the action branch with the maximum upper bound value:
\begin{equation}
\optact=\argmax_{\act \in \actset} \uba{\node}{\act}
\label{Eqn:select_ub}
\end{equation}
and select the observation branch leading to a child node \newnode with the maximum weighted excess uncertainty (WEU): 
\begin{eqnarray}
\optobs&=&\argmax_{\obs \in Z_{\node,{\optact}}} \weu{\newnode}\\
&=&\argmax_{\obs \in Z_{\node,{\optact}}} \left\{\gap{\newnode}-\frac{\scenariosetsize{\newnode}}{\nscen}\cdot \xi \gap{\rootnode}\right\}
\label{Eqn:select_lb}
\end{eqnarray}
Here $\scenariosetnode{\newnode}$ denotes the set of scenarios visiting \newnode, and $\gap{\node}=\ub{\node}-\lb{\node}$ represents the gap between the upper and lower bound in node \node. Intuitively, the WEU value \weu{\newnode} captures the amount of uncertainty contained in node $\newnode$ with reference to that in the root node \rootnode. DESPOT terminates an exploration path if $E$ become zero at the current node. The constant $\xi$ controls the target level of uncertainty to be achieved by the search. 

\begin{figure*}[!t]
  \centering
  \includegraphics[width=0.8\textwidth]{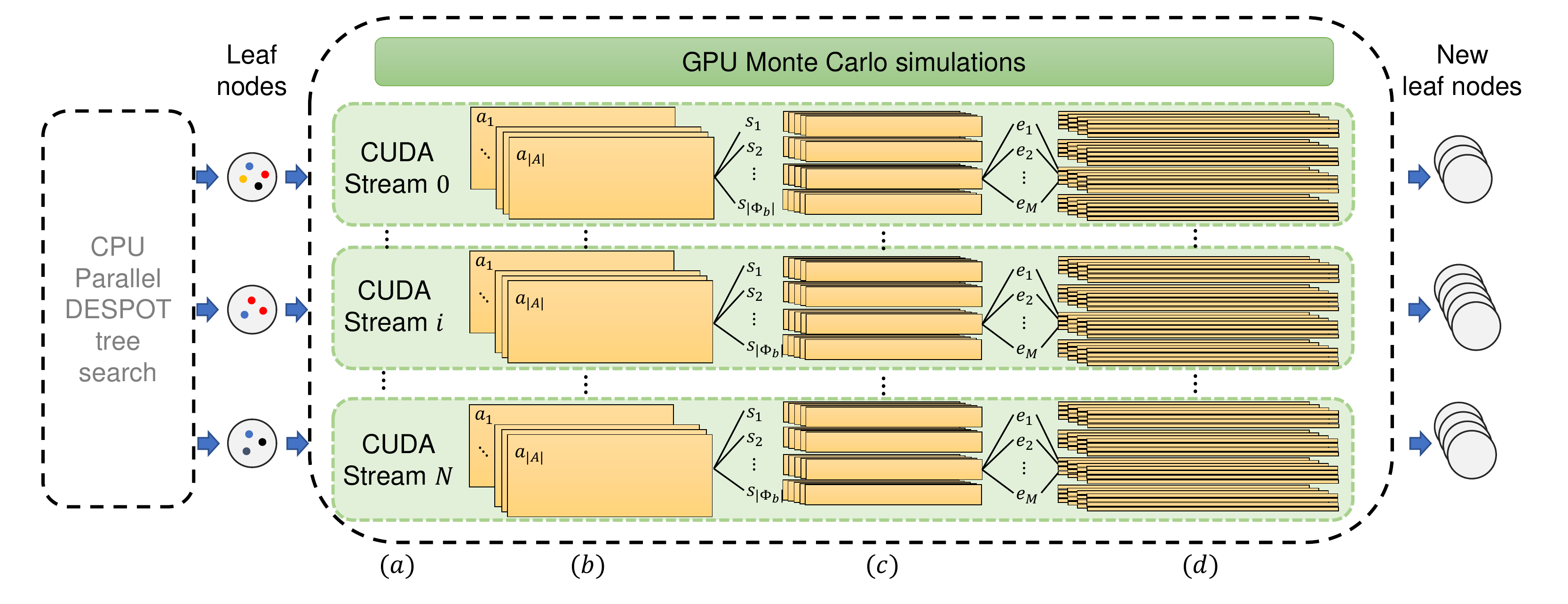}
  \caption{Multi-level parallelization scheme for Monte Carlo simulations in
    HyP-DESPOT. (a) Node-level parallelism. (b) Action-level parallelism. (c)
    Scenario-level parallelism. (d) Fine-grained \elementlevel parallelism.}
  \label{multi-level-parallelism}
\end{figure*}

\subsubsection{Scenario-based PO-UCT for Action Branches}
The PO-UCT algorithm \cite{POMCP} is originally designed for trading off exploitation and exploration during the serial belief tree search. 
It augments the value of an action branch with an exploration bonus that captures its frequency of been tried, such that the search not only exploits the known promising directions, but also reduces the uncertainty in less-explored branches. 

We reformulate the PO-UCT algorithm to distribute parallel CPU threads across action branches under HyP-DESPOT nodes.
The new algorithm, \emph{scenario-based PO-UCT}, respects that a belief node \node in DESPOT is always traversed by a set of scenarios, \scenariosetnode{\node}. 
It records a scenario-wise visitation count for each node \node, written as $\scenariosetsize{\node}\nvisit{\node}$, and for each action branch under \node , written as $\scenariosetsize{\node}\nvisita{\node}{\act}$, then uses the following augmented upper bound to select an action branch for a thread:

\begin{equation} \label{equation::UCT}
\augub{\node}{\act}=\uba{\node}{\act}+\ascale\sqrt{\frac{log(\scenariosetsize{\node}\nvisit{\node})}{\scenariosetsize{\node}\nvisita{\node}{\act}}}
\end{equation}
The last term in Eqn. (\ref{equation::UCT}) is the exploration bonus, which is updated immediately when each CPU thread visits \node. The scaling factor \ascale controls the desired level of exploration for CPU threads among action branches, and can be tuned offline using hyper-parameter selection algorithms like Bayes optimization \cite{bayesopt}.
Note that when using a single scenario ($\scenariosetsize{\rootnode}=1$) in HyP-DESPOT, this exploration bonus is equivalent to that in PO-UCT, but in a parallel setting. 

\subsubsection{Virtual Loss for Observation Branches}
Observation branches under an action captures possible outcomes of the action under different scenarios. It is beneficial to have CPU threads explore multiple of them simultaneously. 
To achieve this, HyP-DESPOT appends a virtual loss \vloss to the WEU value of an observation branch, once it is being traversed by a thread:
\begin{equation} \label{equation::augmented_WEU}
\augweu{\newnode}=\weu{\newnode}-\vloss(\newnode)
\end{equation}
This virtual loss discourages following threads to traverse the same branch, until the current thread leaves the branch and releases it.

In effect, the first thread will always traverse the maximum-WEU observation branch, while later peer threads tend to explore other promising branches. As a simple implementation, $\vloss(\newnode)$ can be set proportional to the initial gap of the root node, written as $\oscale\gap{\rootnode}$, where \oscale controls the level of exploration among observation branches and can be also tuned offline.

\subsection{Parallel Monte Carlo Simulation}\label{leaf_expansion_initialization}

During the search, HyP-DESPOT uses the GPU to continually take over leaf nodes, and perform parallel Monte Carlo simulations to expand them and initialize their children. 

Multiple leaf nodes may be expanded simultaneously. HyP-DESPOT expands each leaf node \leafnode for one level forward, by simulating all possible actions in \actset and all scenarios in $\scenariosetnode{\node}$ \emph{in parallel}, using the deterministic step function:
\begin{equation}\label{eqn:expand}
\snext,\obs=\dsmodelf{\scurr}{\act}{\scenario}, \forall \scenario~ \in \scenariosetnode{\node},~ \act \in \actset 
\end{equation}
We then calculate \emph{in parallel} the initial upper bound and lower bound values for all children belief nodes $\{\newnode\}$:
\begin{equation}\label{eqn:newnodes}
\newnode=\beliefupdate(\belief,\act,\obs),~\act\in\actset,\obs\in {\obsset_{\scenariosetnode{\node},\act}}
\end{equation}
The upper bound is calculated using a heuristic function \ub{\scenario}, and the lower bound is calculated by simulating a default policy \defaultpol from the current depth $\depth_{\newnode}$:
\begin{eqnarray}
\iniub{\newnode}&=&\frac{1}{\scenariosetsize{\newnode}}\sum \limits_{\scenario \in \scenariosetnode{\newnode}} \ub{\scenario}\label{eqn:ub}\\
\inilb{\newnode}&=&\frac{1}{\scenariosetsize{\newnode}}\sum_{\scenario \in \scenariosetnode{\newnode}}\sum\limits_{\timestep=\depth_{\newnode}}^{\infty} \gamma^{\timestep-\depth_{\newnode}} \rfun{\scurr_{\scenario}^{\timestep}(\defaultpol)}{\act^\timestep_{\defaultpol}}\label{eqn:lb}
\end{eqnarray}
where $\scurr_{\scenario}^\timestep(\defaultpol)$ represents the state at time step \timestep, updated using the step function \dsmodel, and determined by scenario \scenario and the sequence of actions $\{\act^\timestep_{\defaultpol}\}$. In practice, we only perform the simulation until a maximum depth \maxsearchdepth, after which the future value is estimated by a heuristic function \lb{\scenario}.

HyP-DESPOT parallelizes all computations in Eqn. (\ref{eqn:expand}), (\ref{eqn:ub}) and (\ref{eqn:lb}) in the GPU, but creates the new nodes (Eqn. (\ref{eqn:newnodes})) in the CPU. 

Modern GPUs has a hierarchical computational architecture, CUDA \cite{CUDA}. GPU functions are launched as ``kernels'' and are executed by a pool of parallel GPU threads. 
The thread pool is organized into multiple thread blocks that are further partitioned into ``warps'' of 32 threads that execute in lock-step. 

Following this architecture, we also parallelize the Monte Carlo simulations in hierarchical levels (\figref{multi-level-parallelism}), including the node-, action-, scenario-, and \element- level. 
The node-level parallelism handles concurrently multiple leaf nodes discovered by CPU threads. The action-level and scenario-level parallelism perform Monte Carlo simulations for different expansion actions and scenarios simultaneously. Finally, the \elementlevel parallelism parallelizes the factored dynamics or observation models (if available) within a simulation step \dsmodel in a fine-grained level.

\subsubsection{Node-level Parallelism and Kernel Concurrency}
When a CPU thread reaches a leaf node, HyP-DESPOT launches a GPU kernel, \emph{MC\_simulation}, to perform the computations defined in Eqn. (\ref{eqn:expand}), (\ref{eqn:ub}) and (\ref{eqn:lb}). HyP-DESPOT associates each CPU thread with a CUDA stream \cite{CUDA}, such that \emph{MC\_simulation} kernels for leaf nodes execute independently and concurrently in the GPU. \figref{multi-level-parallelism}(a) shows the node-level parallelism.
    
\subsubsection{Action-level and Scenario-level Parallelisms}
The \emph{MC\_simulation} kernel for a leaf node \leafnode performs several tasks---\emph{update}, \emph{expansion}, and \emph{roll-out}---using expansion actions in \actset and scenarios in $\scenariosetnode{\node}$. 
Expansion actions are assigned to thread blocks in the GPU (\figref{multi-level-parallelism}(b)), while independent scenarios are assigned to individual threads (\figref{multi-level-parallelism}(c)).
Initially, the leaf node \leafnode to be expanded only contains a set of indexes with respect to the scenario list in its parent node. The \emph{MC\_simulation} kernel first gathers the required scenarios from the parent, and \emph{updates} the scenarios to the current search depth by applying the last action in the history.
The kernel then performs one-level full \emph{expansion} for the leaf node according to Eqn. (\ref{eqn:expand}),
producing stepped scenarios, rewards, and observation labels for all its children nodes.
The kernel further calculates upper bounds for the new nodes using Eqn. (\ref{eqn:ub}), and performs \emph{roll-outs} to initialize the lower bounds using Eqn. (\ref{eqn:lb}).
Then, the \emph{MC\_simulation} kernel returns the scenario-based observations, step rewards, and initial bounds back to the host memory.
The corresponding CPU thread prepares new action branches, observation branches, and children leaf nodes, according to Eqn. (\ref{eqn:newnodes}). 
Finally, the corresponding CPU thread resumes back to the tree search process.

\subsubsection{\Elementlevel Parallelism}
For large-scale problems, the dynamics or observation models in the step function \dsmodel often have multiple independent elements. 
For example, a problem may have multiple robots or dynamic objects in the environment moving independently.
We can thus factor the models into fine-grained parallel tasks (\figref{multi-level-parallelism}(d)). 
HyP-DESPOT assigns these tasks to different thread warps in the GPU, to avoid potential serialization problem caused by heterogeneous tasks, \eg, transitions of a vehicle and transitions of pedestrians in a car driving problem. 
By applying this within-step factorization, HyP-DESPOT achieves an increased level of parallelism and thus higher GPU utilization. It also helps to reduce the memory usages in GPU blocks, because each block needs to process less scenarios.

\section{Experimental Results}

We evaluated HyP-DESPOT in simulation on three large-scale planning tasks
under uncertainty: navigation with a partially known map, multi-agent rock
sample, and autonomous driving in a crowd. The navigation task has an enormous
state space of size $|\stateset|=169\times 2^{124}$, because of map
uncertainty.
The multi-agent rock sample task has $625$ actions, requiring HyP-DESPOT to
search a very large tree.  Finally, the autonomous driving task has a huge
observation space with more than $10^{112}$ observations and a complex
dynamics model, and, we evaluated HyP-DESPOT both in simulation and on a real
robot vehicle.  We compare HyP-DESPOT with the original DESPOT algorithm and
GPU-DESPOT, which performs GPU parallelization only. Our results show that
HyP-DESPOT speeds up DESPOT by up to several hundred times. GPU
parallelization provides significant performance gain, and integration with
CPU parallelization provides additional benefits.

The performance benefits of HyP-DESPOT depends on the inherent parallelism
that a task affords. Our results suggest that generally,  large state and
action spaces have a positive effect on parallelization and large observation
space has  a  negative effect.

Details are presented in the subsections below.

\subsection{Evaluation Tasks}
\begin{figure}[!t]
  \centering
  \begin{tabular}{c@{}c@{}c}
   \hspace{-0.3cm}\includegraphics[height=1in]{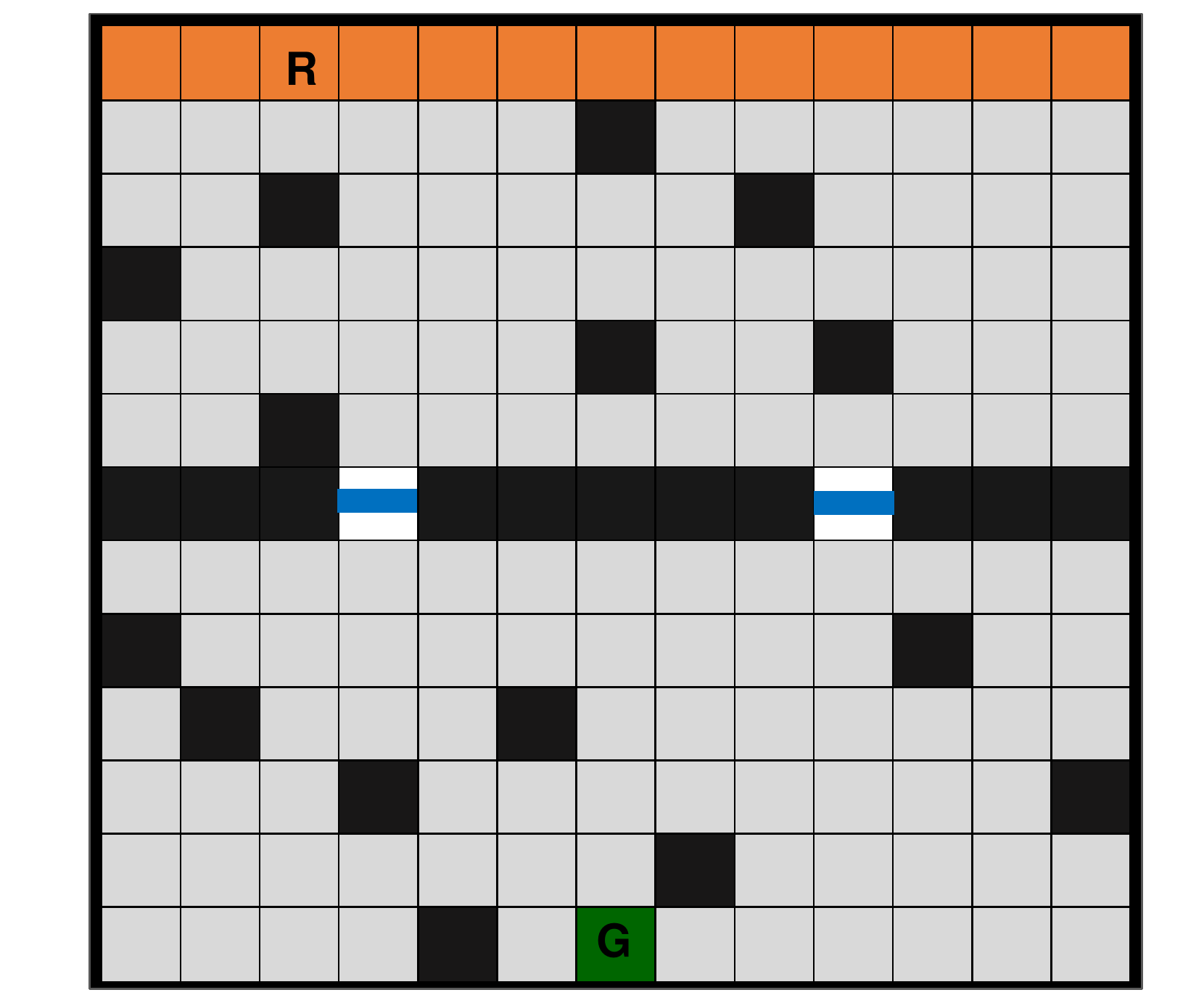}
   & \includegraphics[height=1in]{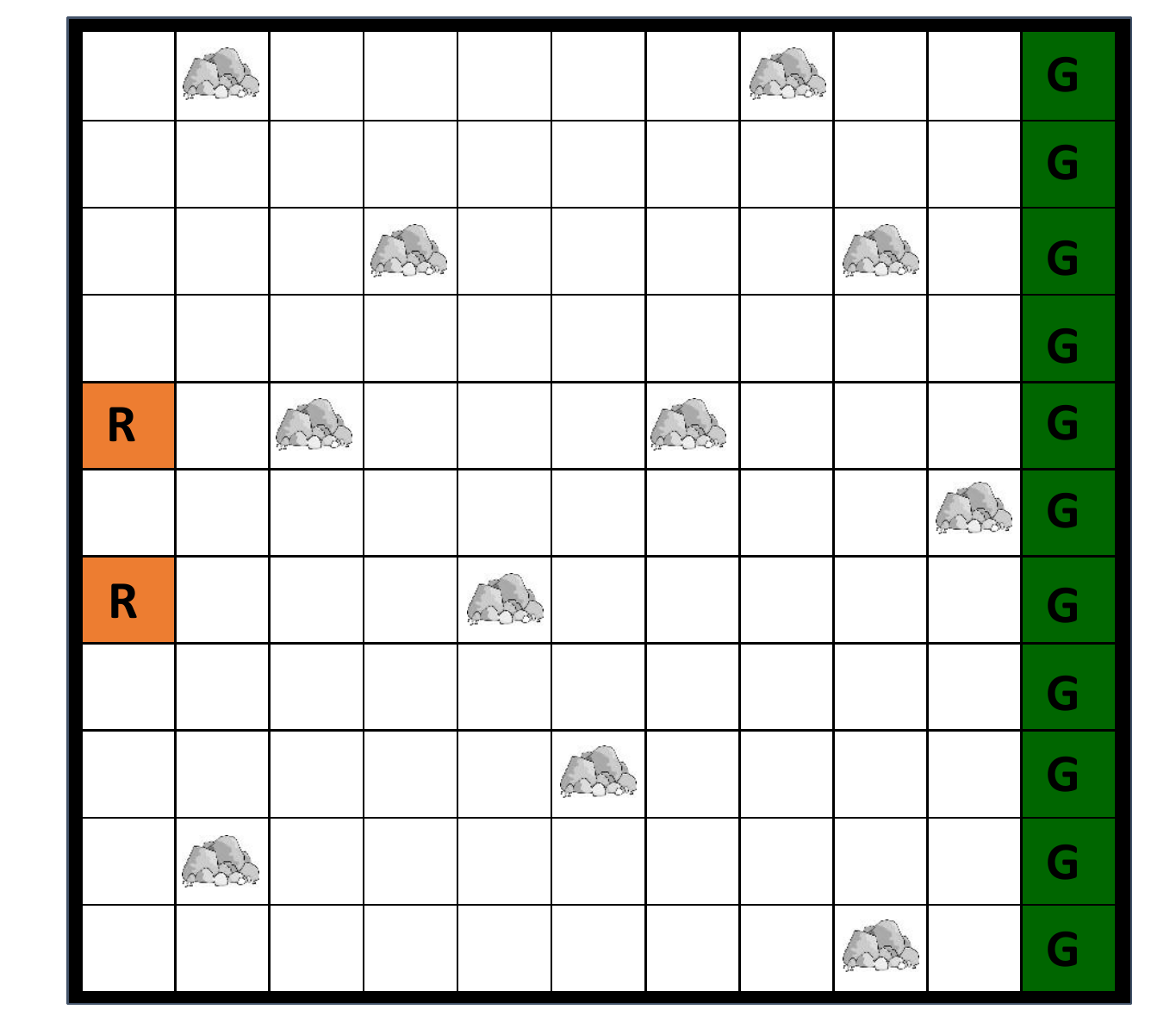}
   & \includegraphics[height=1in]{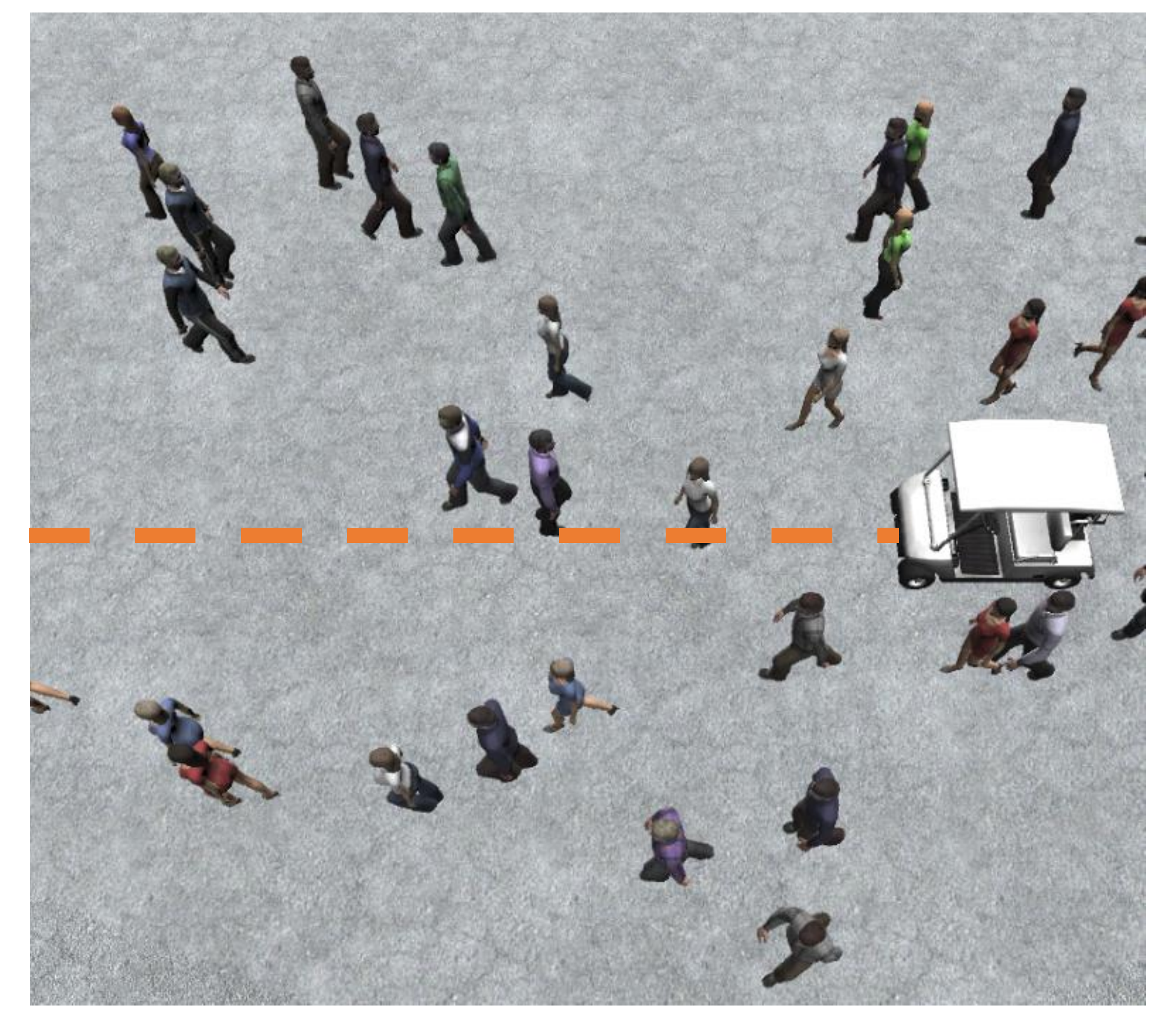} \\
   (a) & (b) & (c)
  \end{tabular}
  \caption{Large-scale planning tasks for evaluating HyP-DESPOT. (a)
    Navigation with a partially known map. (b) Multi-agent rock sample. (c)
    Autonomous driving in a crowd.}\label{fig:problems}
\end{figure}

\subsubsection{Navigation in a Partially-Known Map}\label{Section::Nav}

In this task (\figref{fig:problems}(a)), a robot starts from a random position at the top border of a $13\times13$ map, and travels to its goal in the bottom via one of the two alternatively open gates on the middle wall (colored in blue in \figref{fig:problems}(a)). 
The map is only partially-known to the robot. The known grids (black grids in \figref{fig:problems}(a)) help the robot localize itself, but they look identical to each other. For other grids (grey in \figref{fig:problems}(a)), the robot is only informed that they have 0.1 probability to be occupied. 

In each step, the robot can stay or move to its eight neighboring positions. Both the transition and sensing of the robot are noisy. Moving of the robot can fail with a small probability 0.03, while observations on the eight adjacent grids (\occupied or \free) can also be wrong with 0.03 probability in each direction. Staying still is discouraged by a small penalty (-0.2). The robot receives a small motion cost (-0.1) for each step it moves. If the robot hits an obstacle, it receives a crash penalty (-1). When the goal is reached, the robot receives a goal reward (+20), and the world terminates. 

This navigation task has an huge state space $|\stateset|=169\times 2^{124}$. 
To perform the navigation successfully, the robot has to hedge against both uncertainties in its localization and the shape of the map, and plan for sufficiently long horizon to precisely pass the open gate.

\subsubsection{Multi-agent \RockSample}\label{Section::MARS}
\begin{figure}[!t]
  \centering
  \begin{tabular}{c@{}c@{}c}
  \hspace{-0.2cm}\includegraphics[height=1.3in]{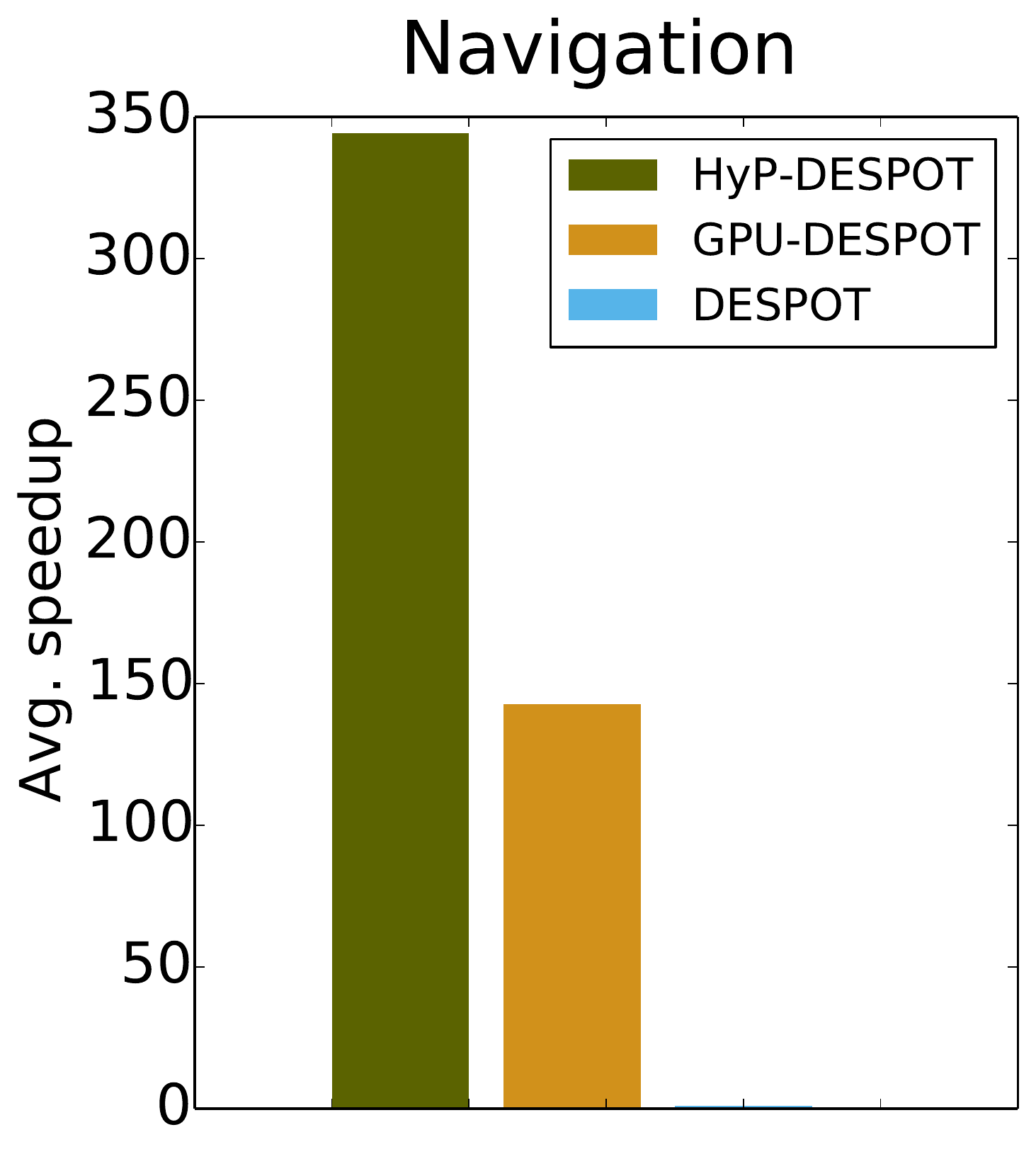}
   & \includegraphics[height=1.3in]{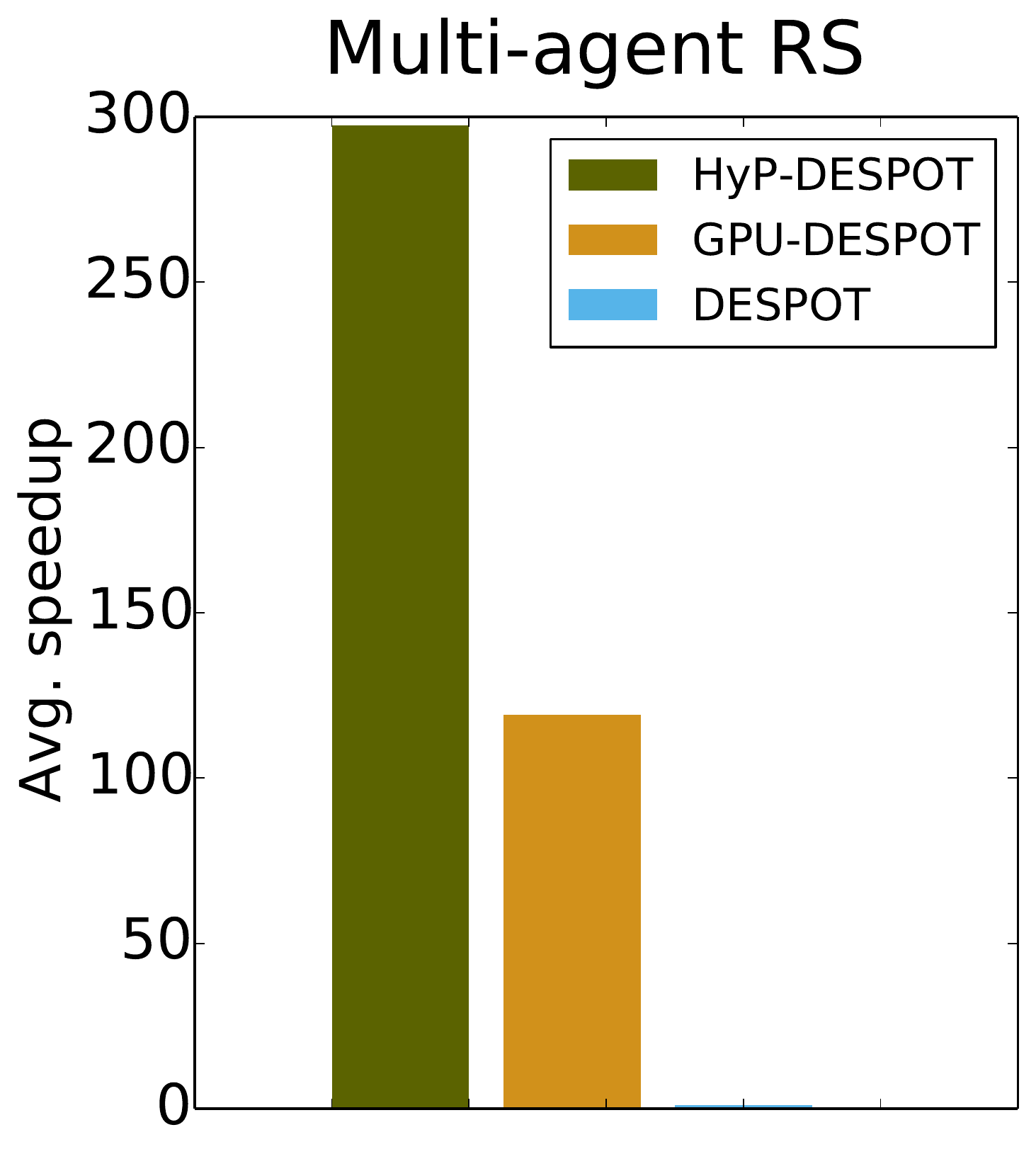}
   & \includegraphics[height=1.3in]{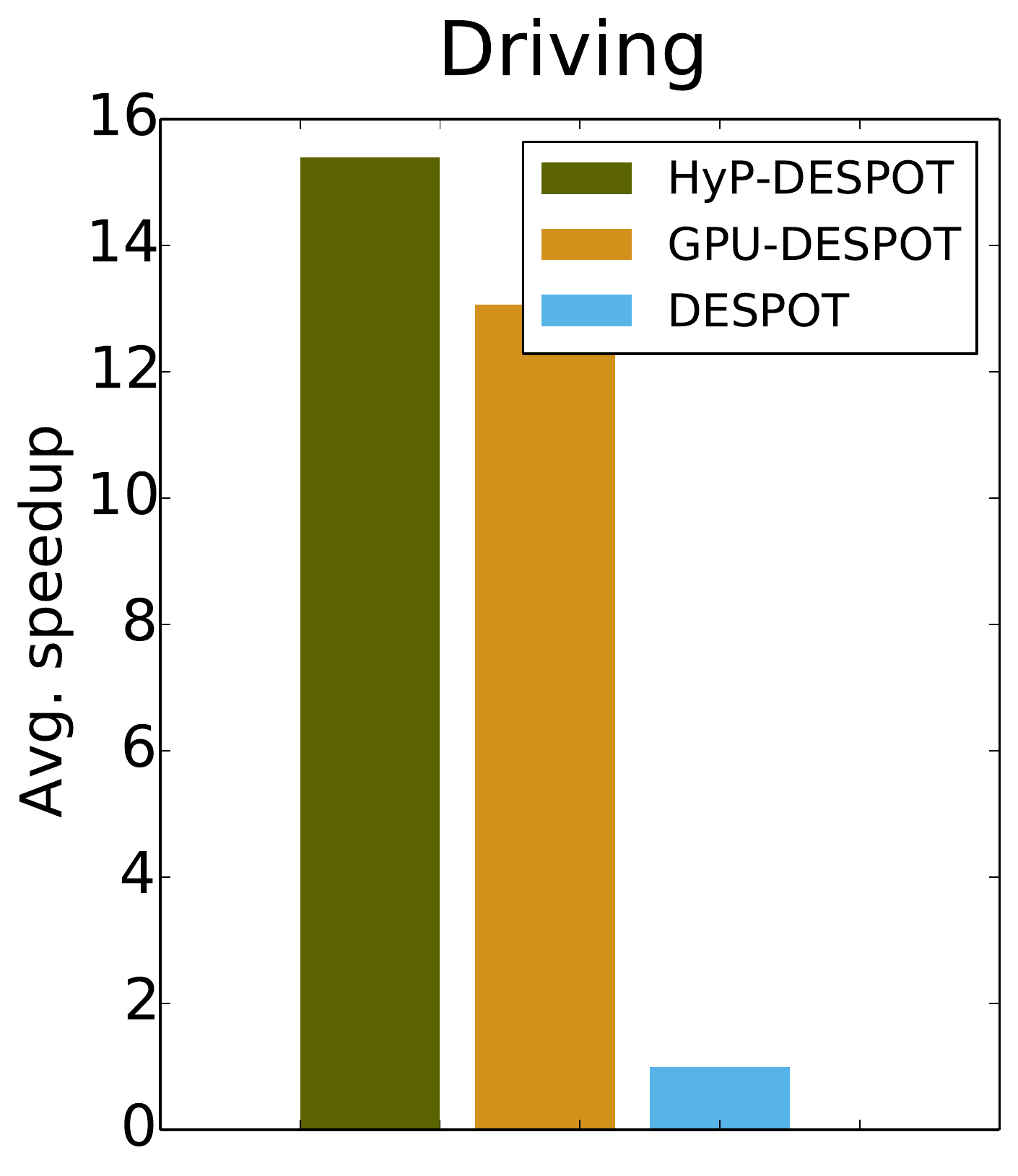} \\
   \hspace{-0.2cm}\includegraphics[height=1.3in]{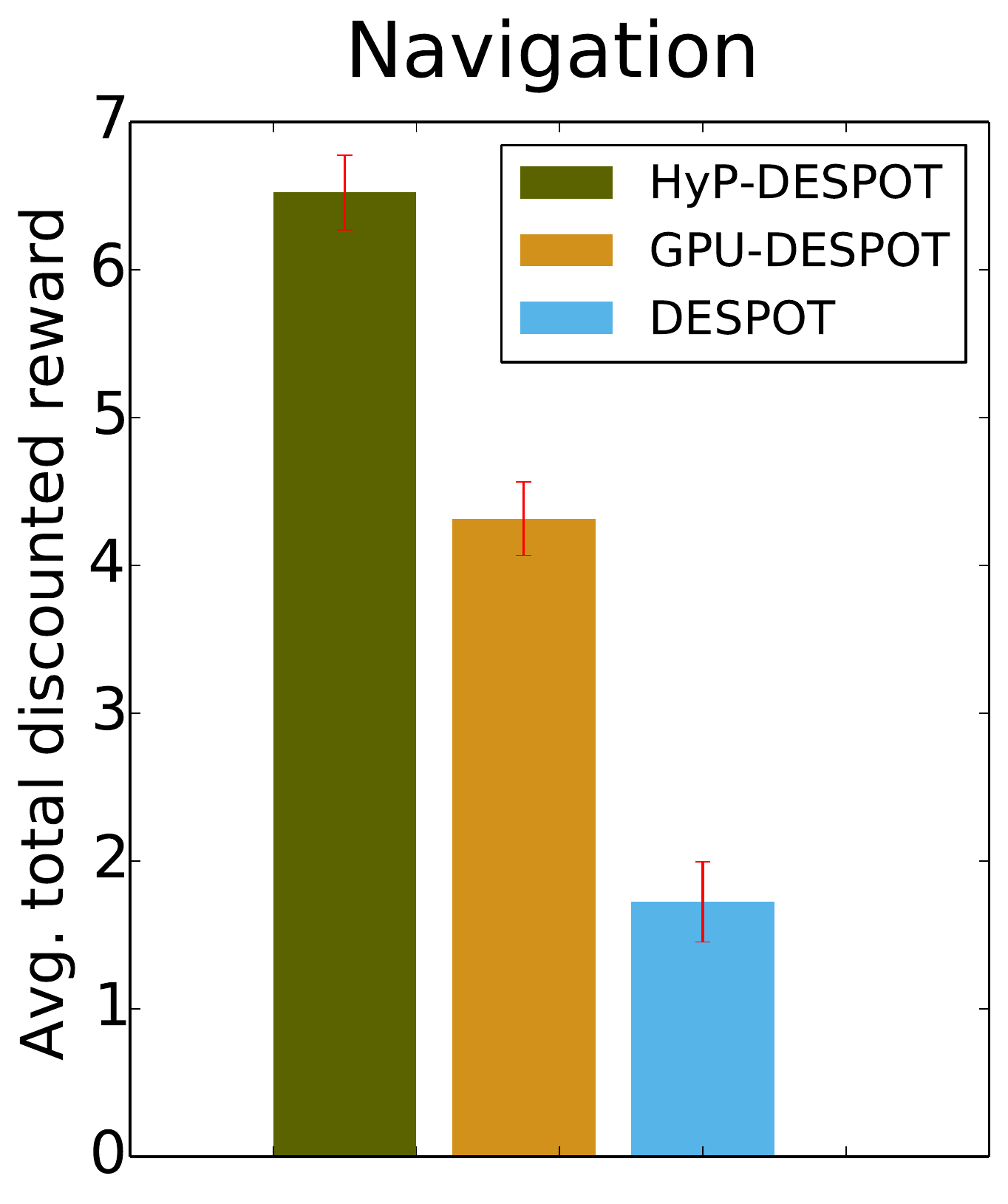}
   & \includegraphics[height=1.3in]{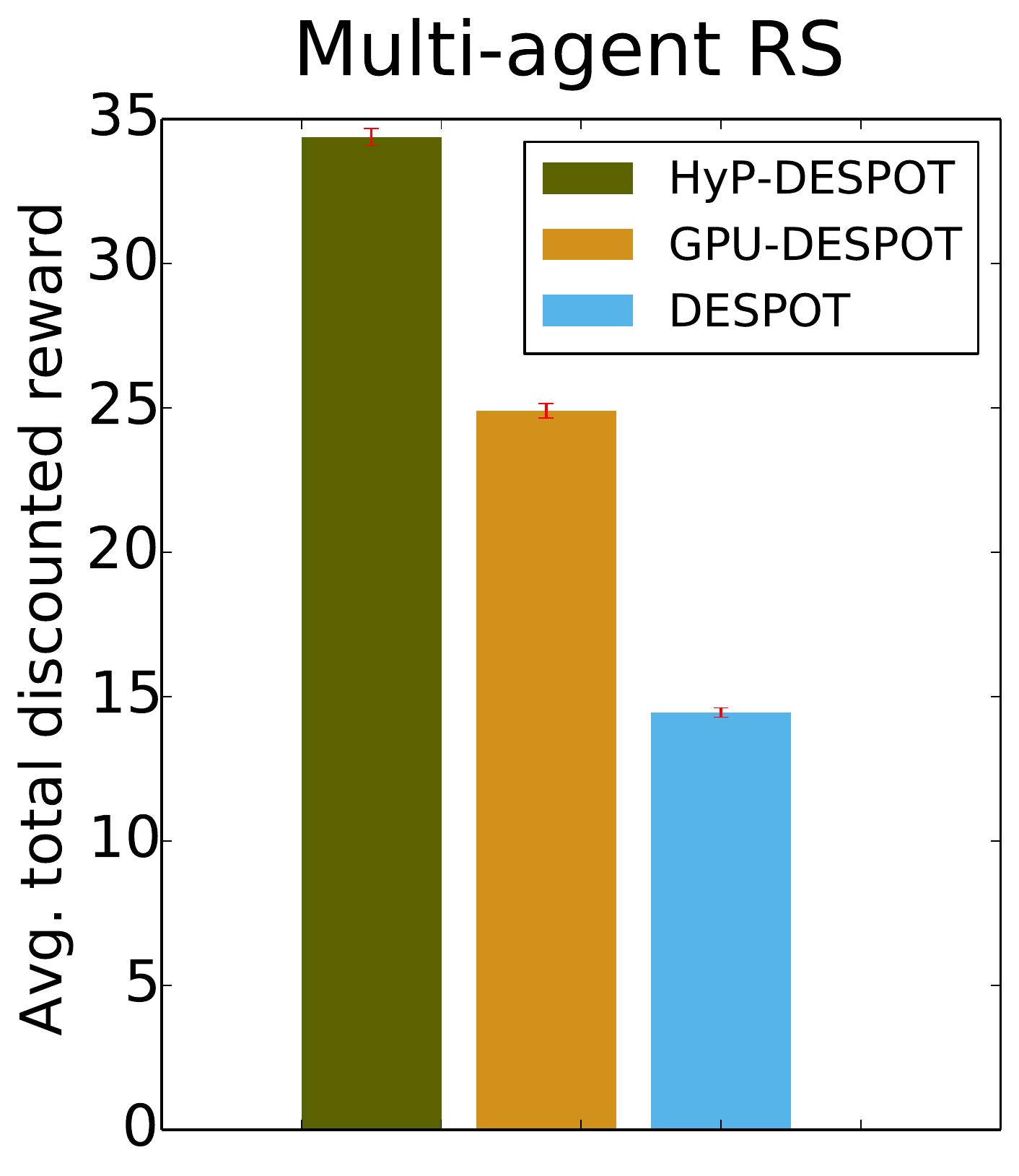}
   & \includegraphics[height=1.3in]{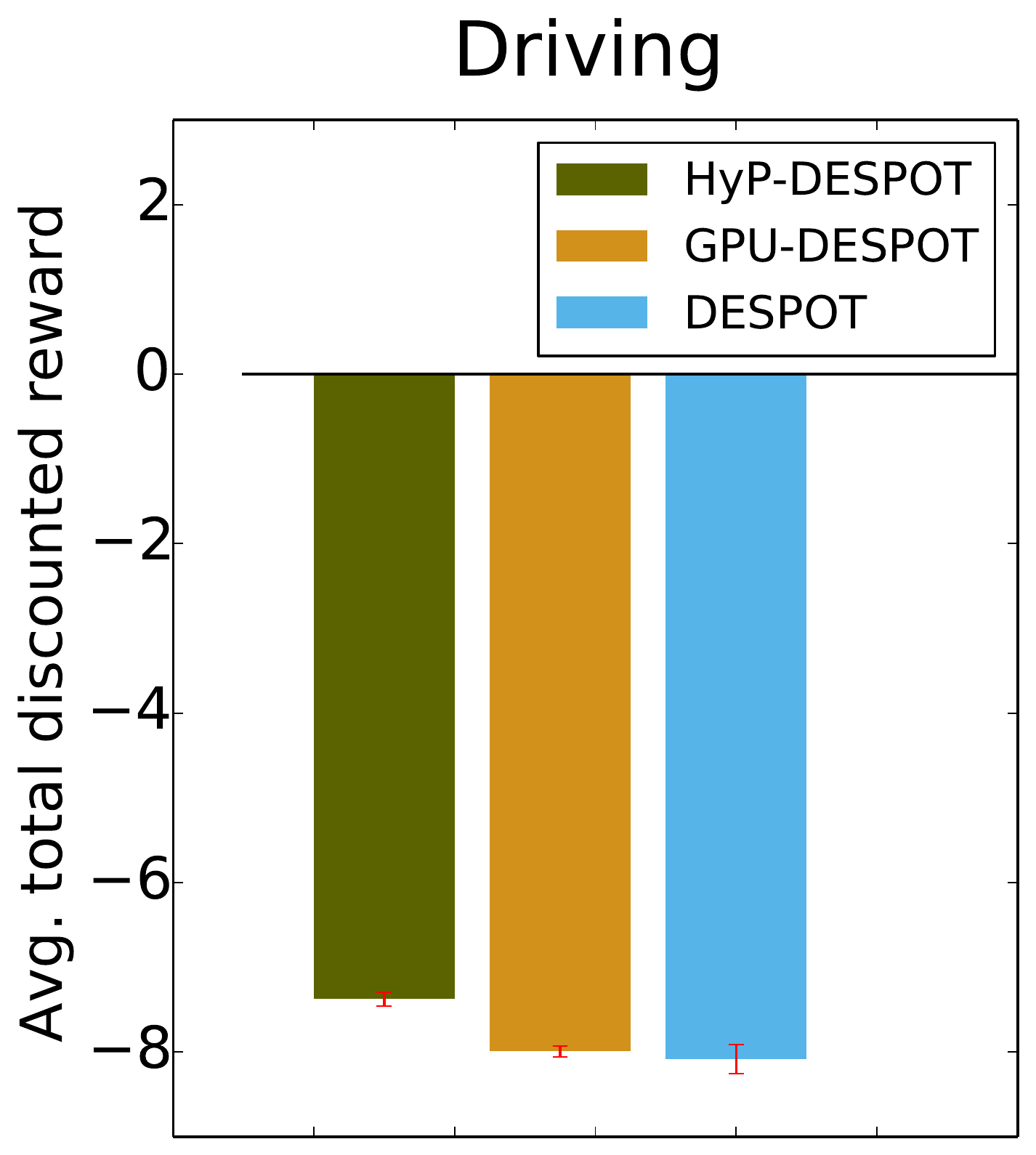} \\
   (a) & (b) & (c)
  \end{tabular}
  \caption{Performance of HyP-DESPOT and GPU-DESPOT, compared with DESPOT, in solving the three evaluation tasks:
  Average speedup (first row),
 and  average total discounted reward (second row).
  }\label{fig:problem_rewards}
\end{figure}

To test the performance of HyP-DESPOT on tasks with many actions, we modify \emph{\RockSample}, a well-established benchmark, to the \emph{multi-agent \RockSample} problem (\figref{fig:problems}(b)). In multi-agent \RockSampleTight$(n,m)$, two robots cooperate to explore a $n\times n$ map and sample $m$ rocks distributed across the map. The robots aim to sample as many \good rocks as possible in total and leave the map via the east border.
The robots are mounted with noisy sensors to detect whether a rock is \good or \bad, with their accuracy decreasing exponentially with the sensing distance. In each step, each robot can choose to either move to the four neighboring grids, or \sense a specific rock. If a robot is already at the same grid with a rock, it can choose to \sample it, and receive a +10 reward if the rock is \good, or a -10 reward if the rock is \bad. Finally, a robot receives a +10 reward upon reaching the east border. The world terminates when both robots exist the map. We test HyP-DESPOT on multi-agent \RockSampleTight(20,20).

Multi-agent \RockSampleTight(20,20) requires centralized planning for the two robots. The task thus has a large action space containing 625 actions, requiring HyP-DESPOT to explore a wide belief tree.

\subsubsection{Autonomous Driving in a Crowd} \label{Section::CarDrving}

\begin{table}[!t]
	\centering
	\vspace{0.8cm}
	\caption{Performance comparisons of  DESPOT (\nscensymbol=100),
          GPU-DESPOT (\nscensymbol=1000), and HyP-DESPOT (\nscensymbol=1000)
          on the autonomous driving task.
        }
	\begin{tabular}{ l p{1.6cm} c c }
		\toprule
		\centering
		&  Collision rate & Traveled distance & Decelerations\\ \midrule
		DESPOT & 0.00177 $\pm~$0.0002 & 12.493 $\pm$ 0.1 & 8.175 $\pm$ 0.07 \\ \midrule 
		GPU-DESPOT& 0.000496 $\pm~$0.00008 & 9.131 $\pm$ 0.07 & 6.744 $\pm$ 0.05 \\ \midrule 
		HyP-DESPOT& 0.000612 $\pm~$0.00008 & 10.034 $\pm$ 0.08 & 6.045 $\pm$ 0.05 \\ 
		\bottomrule
	\end{tabular}
	\label{tab:car_driving}
\end{table}
We also demonstrate the effectiveness and efficiency of HyP-DESPOT with a real-world robotics task: an autonomous vehicle driving through a dense crowd (\figref{fig:problems}(c)). We first conduct a quantitative study in a simulation environment (\figref{fig:problems}(c)), then demonstrate the application on a real robot vehicle in \secref{sec:scooter}.

In this task \cite{Bai_2015}, a vehicle drives along its planed path among a crowd of pedestrians (\figref{fig:problems}(c)), with its speed controlled by a POMDP planner. The vehicle tries to reach its goal within 200 time steps, while taking care of 20 nearest pedestrians. We assume that pedestrians tend to move towards their goals with a uniform speed and Gaussian noises on their heading directions. The vehicle is provided a set of possible goal locations of pedestrians. In each time step, the vehicle can choose to \accel, \decel, or \maintain\xspace its speed, so that it avoids collisions with pedestrians and drives efficiently and smoothly along its path. However, both \accel\xspace and \decel\xspace of the vehicle can fail with a small probability (0.01). Rewards in this task follows the setting in \cite{Bai_2015}.

This task has a huge state space comprising of both observable and hidden variables. We assume that the vehicle can fully observe positions and velocities of itself and all pedestrians around it, but cannot directly know the goals of individual pedestrians, which information has to be inferred from past observations.

\subsection{Performance  Comparison}\label{Section::Performance}

To show the computational efficiency of HyP-DESPOT and GPU-DESPOT over DESPOT, we measure their \emph{speedup}, defined as the size of the belief tree been constructed within a given planning time. Our experimental results show that, building a larger tree leads to higher solution quality, measured by the total discounted reward. If the tree algorithms over-use the planning time when expanding the root node, we further normalize the tree size by the planning time. 

All experiments were conducted
on a server with two Intel(R) Xeon(R) Gold 6126 CPUs running at 2.60GHz, a GeForce GTX 1080Ti GPU (11GB VRAM), and 256 GB main memory. 
The navigation task and multi-agent RS are solved using 1 second planning time, as in standard online planning setting. For the driving task, we use 10 Hz control frequency (0.1 second planning time).



For navigation in a partially-known map, HyP-DESPOT and GPU-DESPOT achieve 344.3 and 142.6 times
speed-up over DESPOT (\figref{fig:problem_rewards}(a)), respectively. As a result, HyP-DESPOT and GPU-DESPOT achieves 278\% and 150\% higher total discounted rewards than DESPOT, respectively.  

For multi-agent \RockSample, HyP-DESPOT and GPU-DESPOT achieve 297.4 times and 119.2 times speedup over DESPOT, and bring up to 137.9\% and 72.3\% of improvements on the total discounted rewards (\figref{fig:problem_rewards}(b)), respectively.

The autonomous driving task affords limited level of parallelism, primarily because of the huge observation space, $\obsset>10^{112}$, causing scenarios to diverge along the exploration paths. 
HyP-DESPOT still achieves 15.4x speed-up and higher quality solutions (\figref{fig:problem_rewards}(c)) over DESPOT, because of both the parallel tree search and the fine-grained \elementlevel parallelism. Detailed measurements in \tabref{tab:car_driving} show that DESPOT drives over-aggressively when using \nscensymbol=100 (it can't handle larger \nscen's in 0.1s). In contrast, HyP-DESPOT significantly reduces the collision rate from DESPOT, and enables the vehicle to driver faster and smoother than GPU-DESPOT. 

\begin{figure}[!t]
\hspace*{0cm}  
\centering
\hspace{-0.0cm}\subfloat{\includegraphics[height=1.4in]{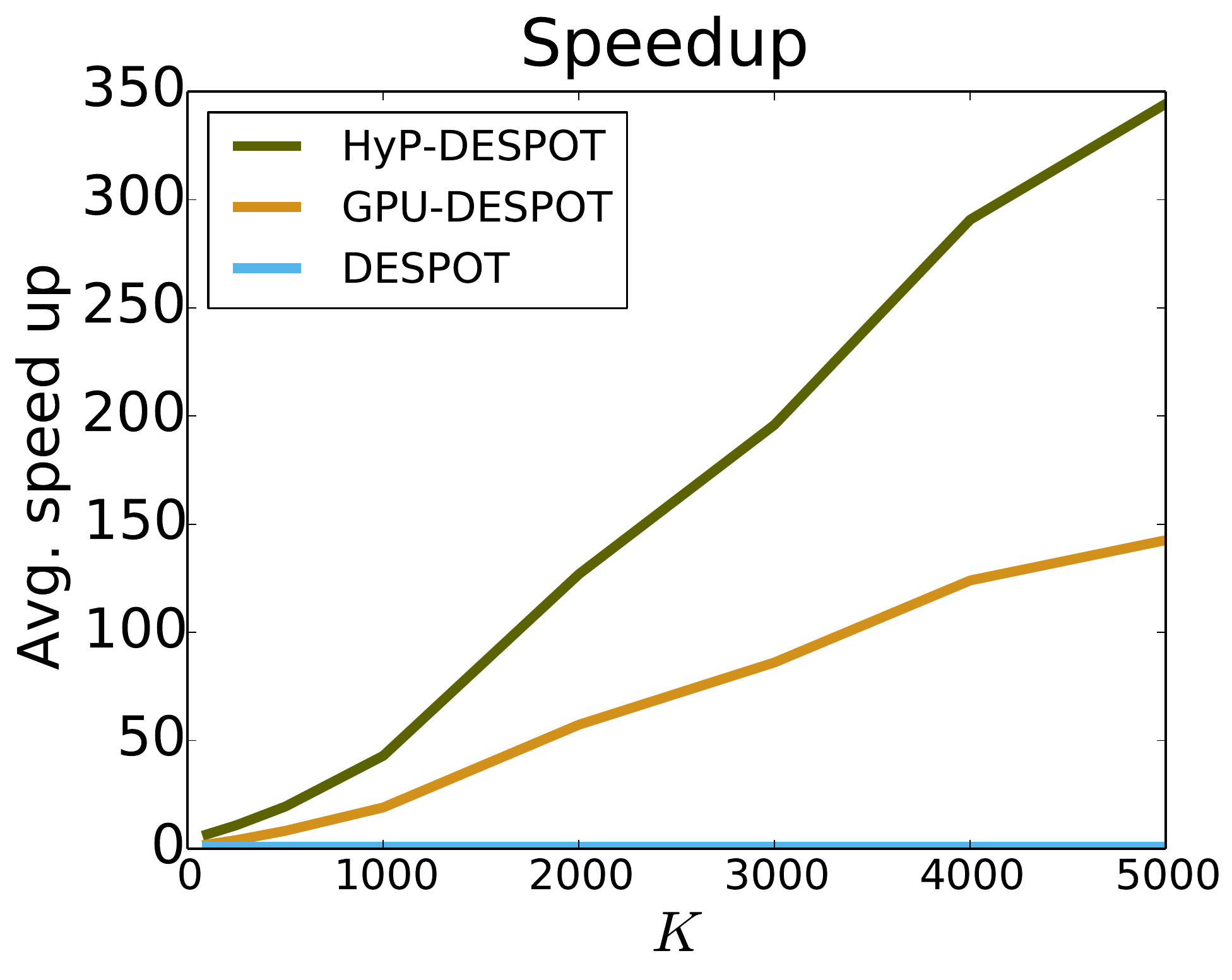}\label{fig:nav_k_vs_speedup}}
\subfloat{\hspace{0.1cm}\includegraphics[height=1.4in]{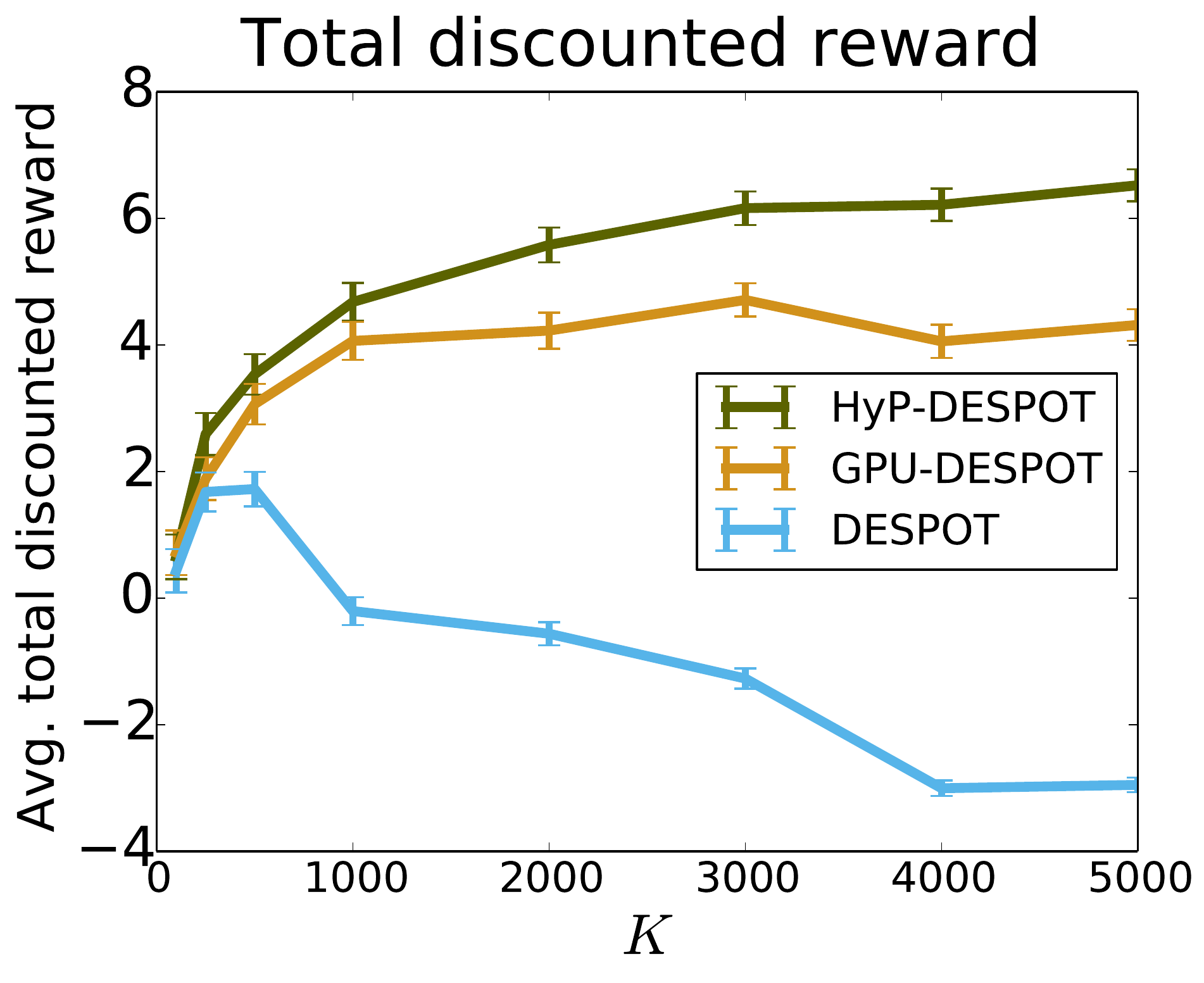}\label{fig:nav_k_vs_rw}}\\
\subfloat{\includegraphics[height=1.35in]{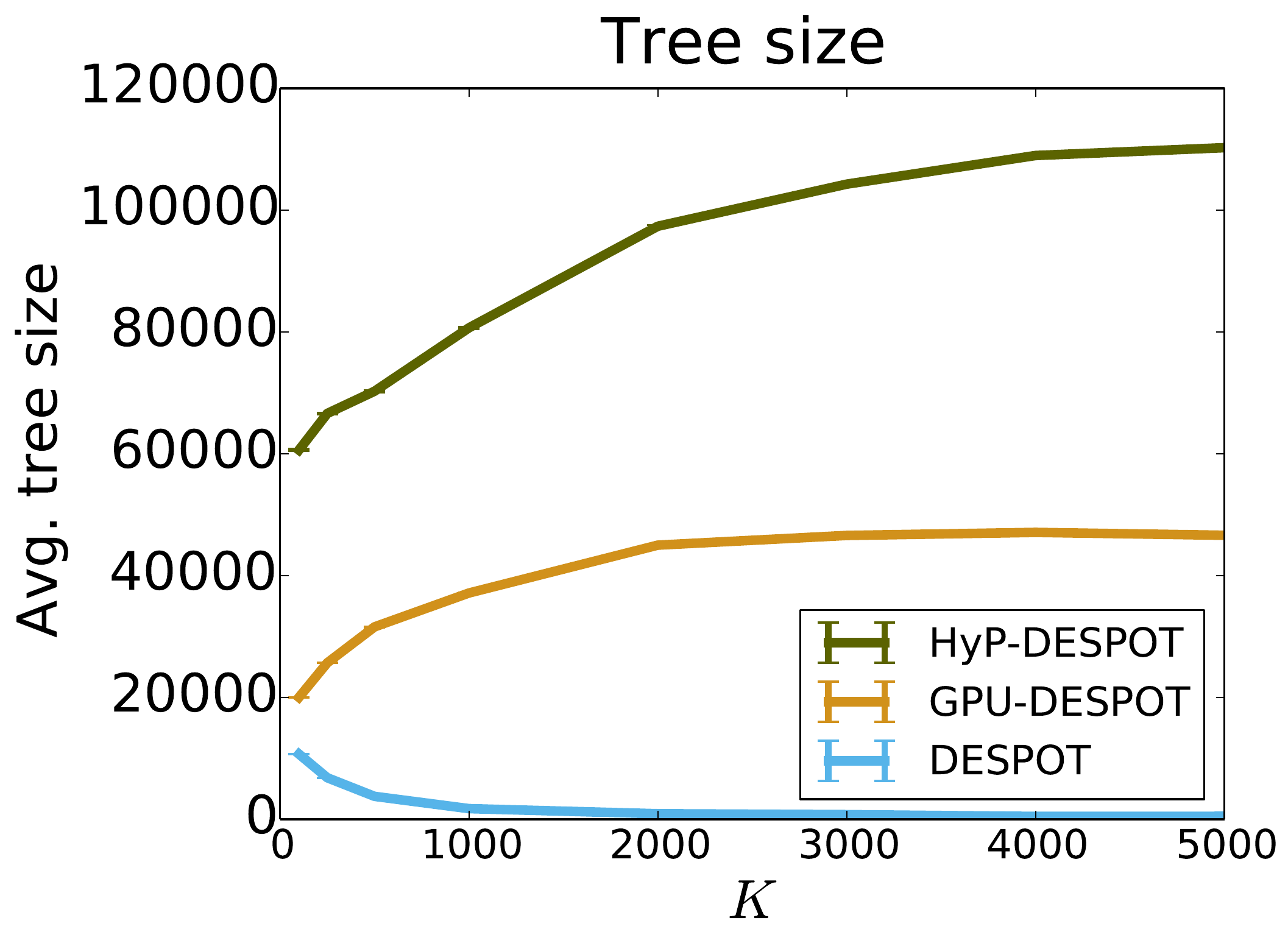}\label{fig:nav_k_vs_tree}} 
\subfloat{\hspace{0.1cm}\includegraphics[height=1.35in]{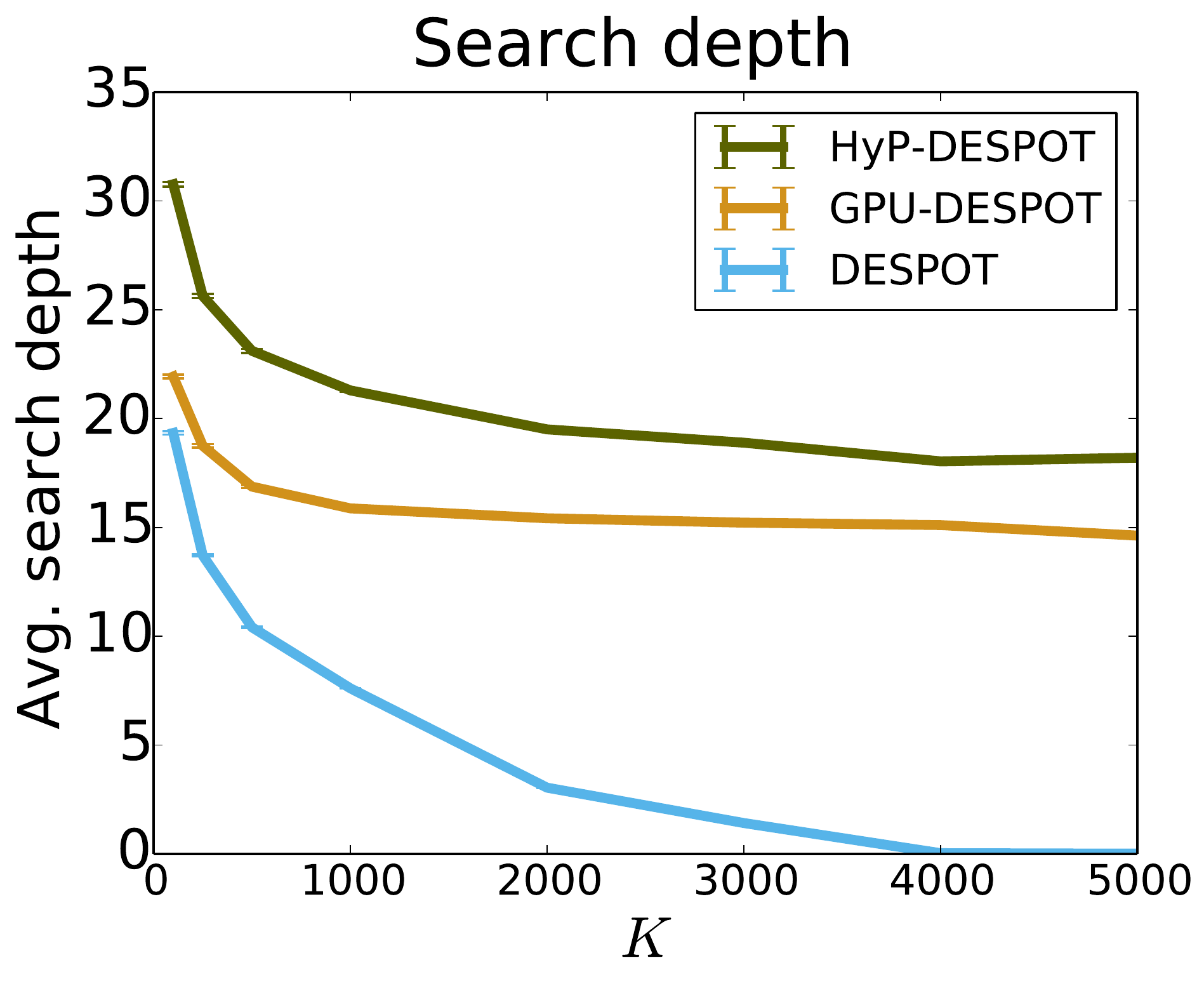}\label{fig:nav_k_vs_depth}}
 \caption{The effect of using different number of scenarios, \nscen, in HyP-DESPOT, for navigation in a partially-known map.}\label{fig:nav_k_vs_all}
\end{figure}

\begin{figure}[!t]
\hspace*{0cm}  
\centering
\subfloat{\includegraphics[height=1.4in]{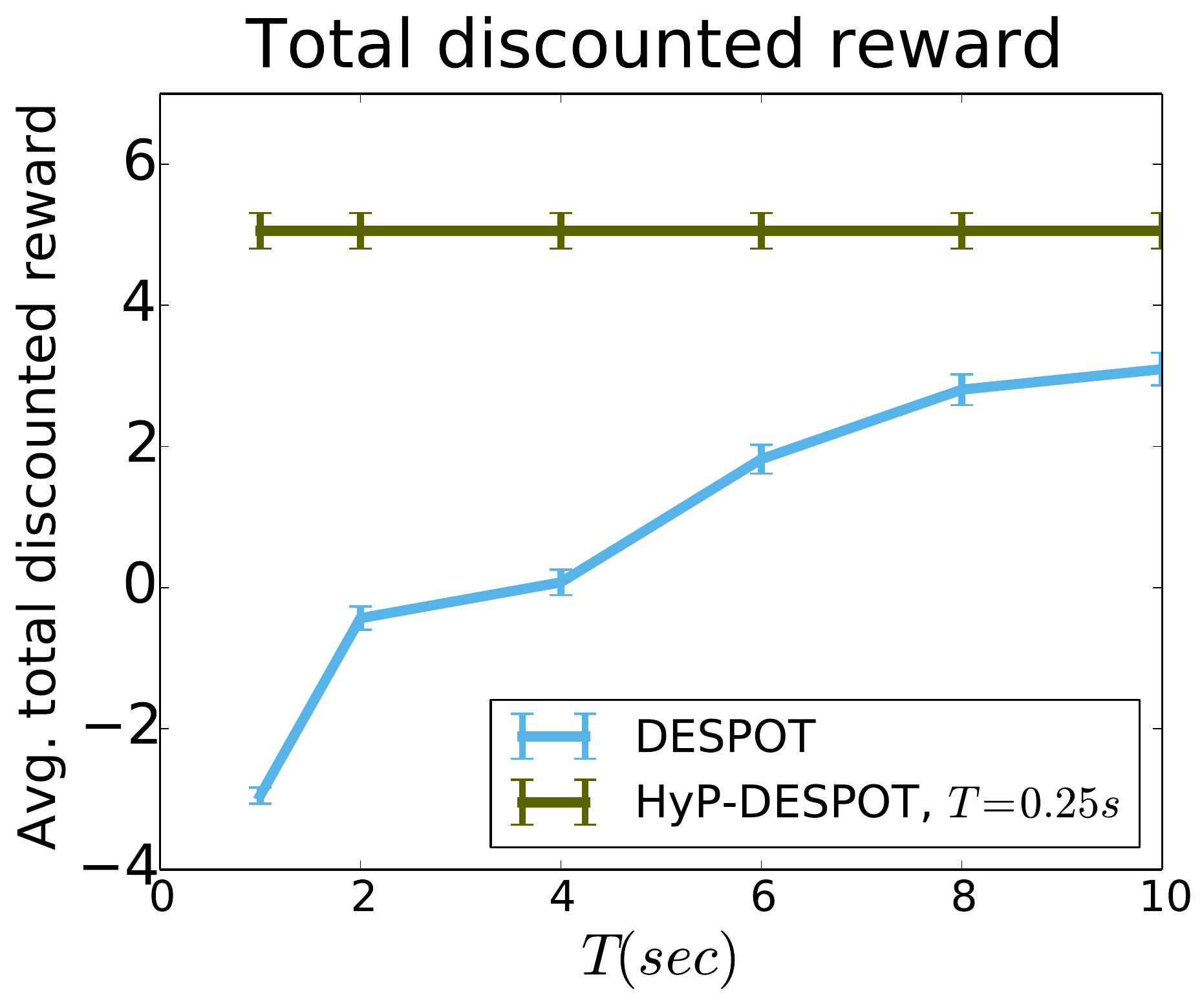}\label{fig:nav_t_vs_rw}} 
\hspace{0.01cm}
\subfloat{\includegraphics[height=1.4in]{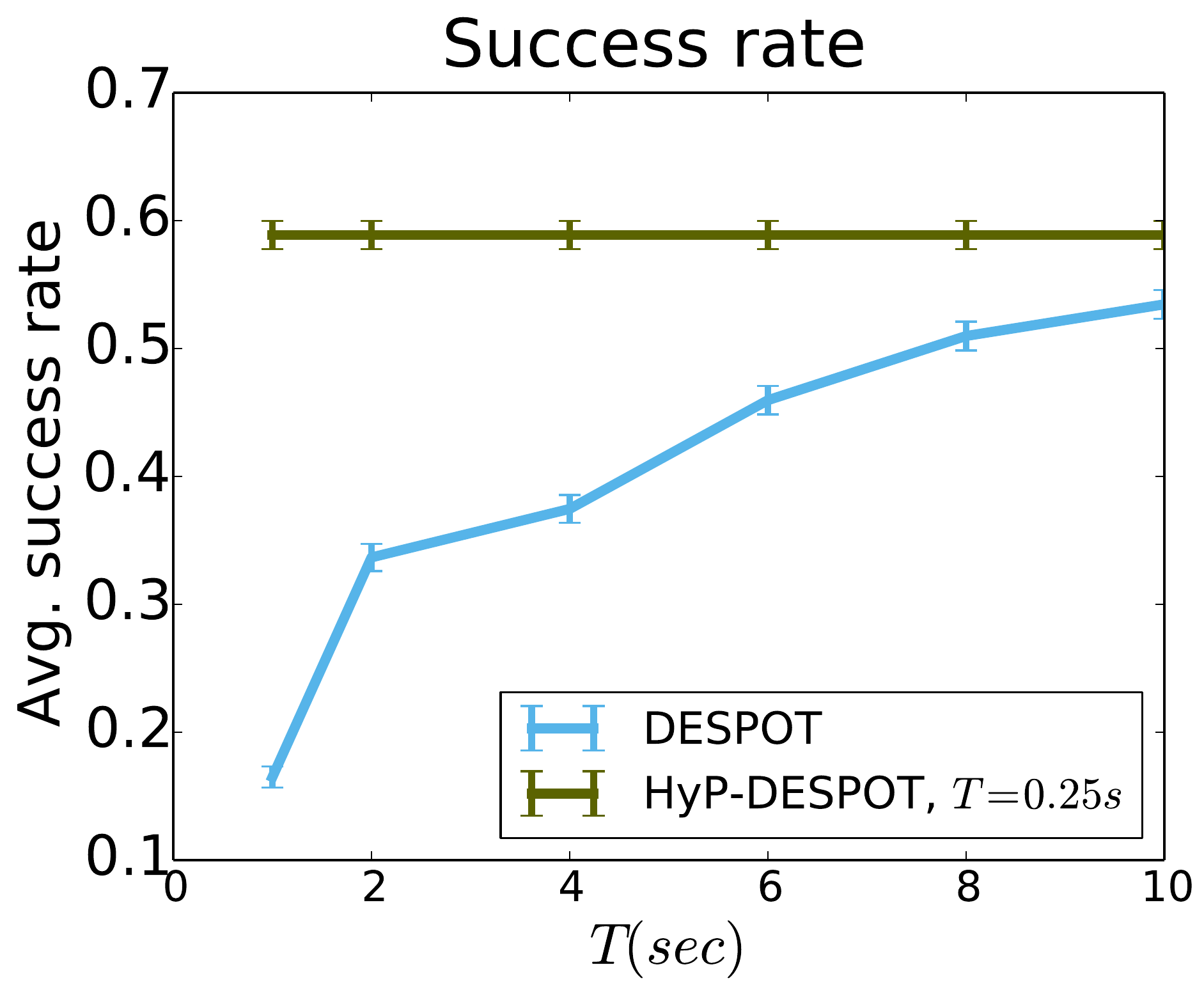}\label{fig:nav_t_vs_succ}}\\
\subfloat{\includegraphics[height=1.39in]{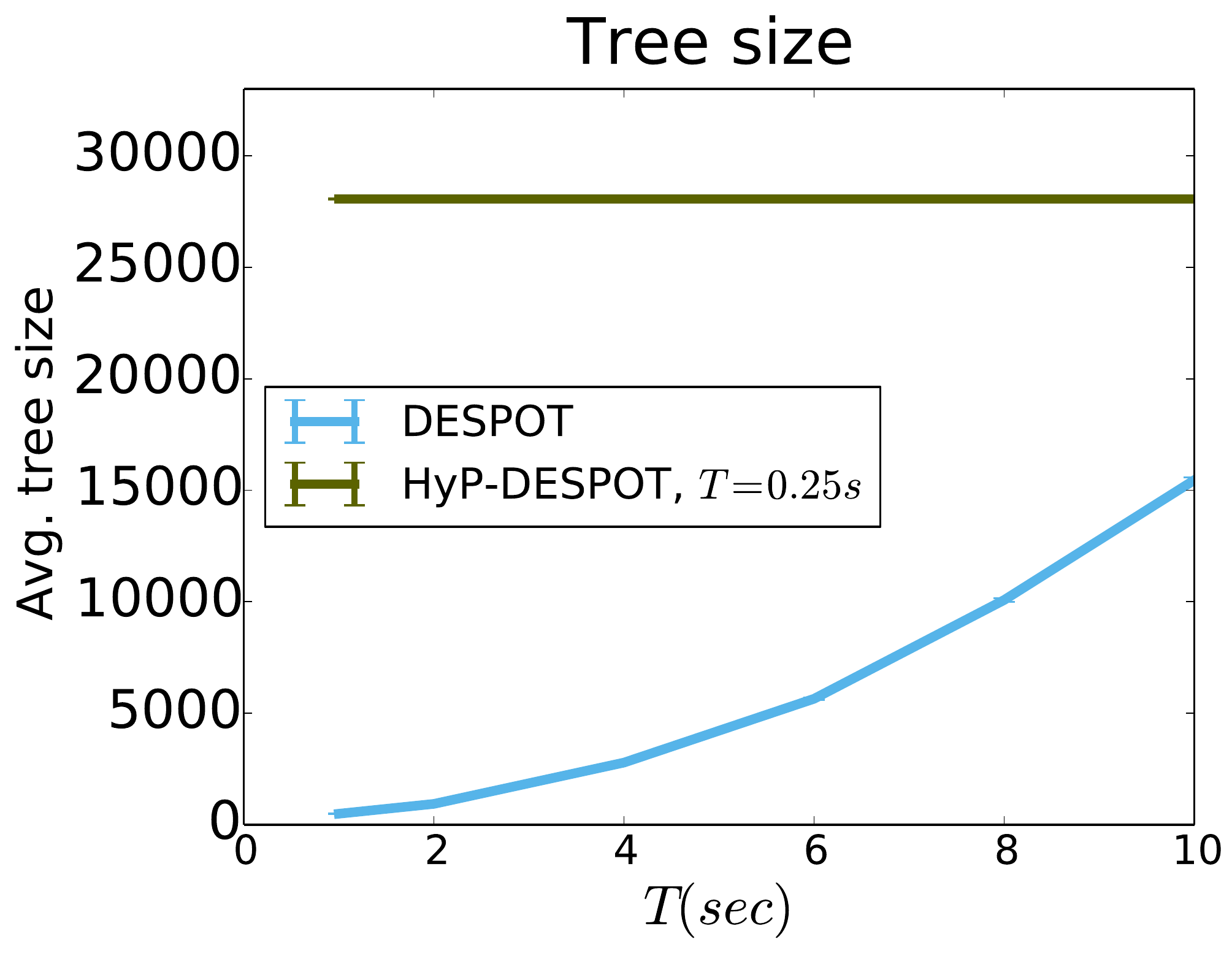}\label{fig:nav_t_vs_tree}} 
\hspace{0.01cm}
\subfloat{\includegraphics[height=1.39in]{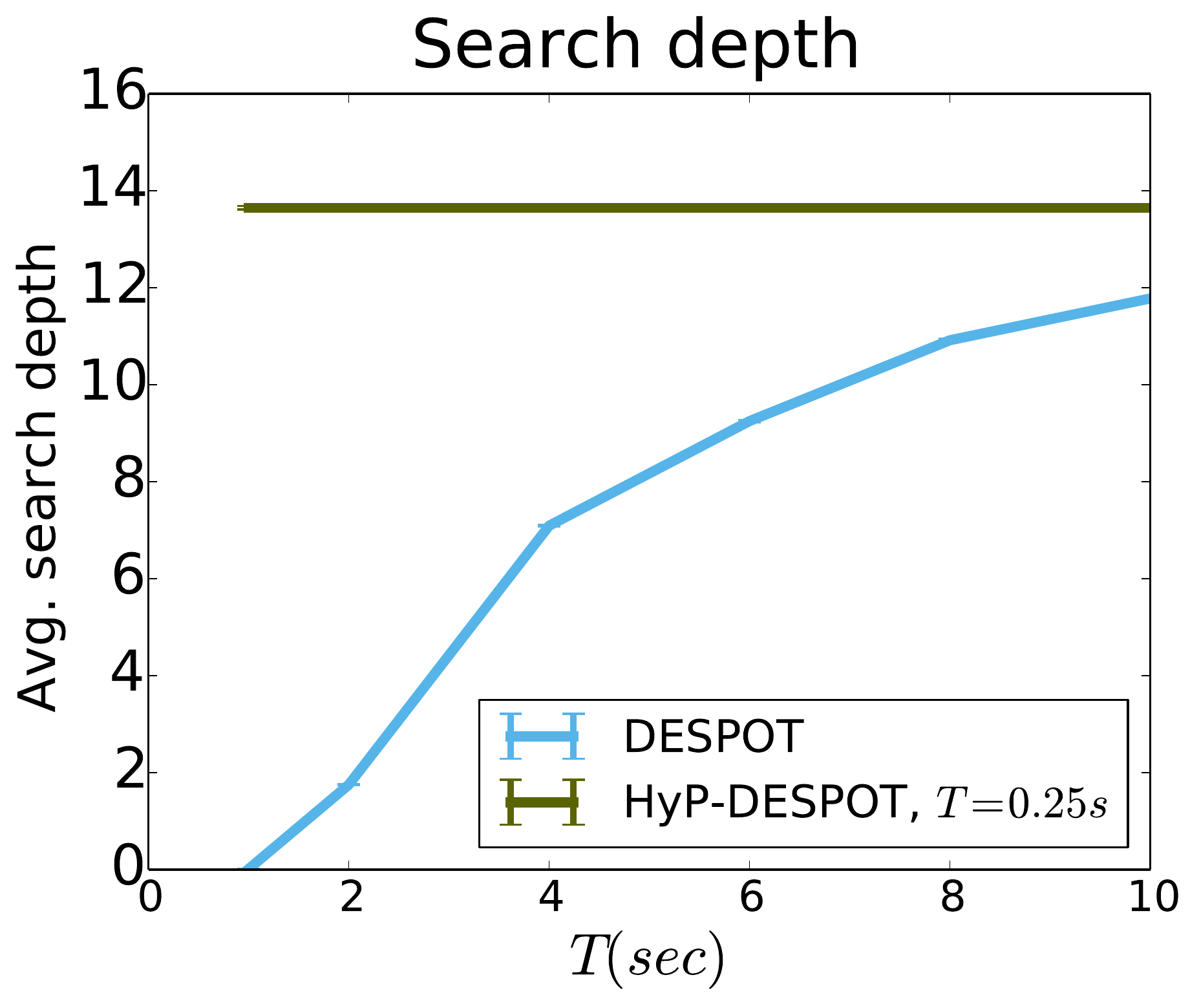}\label{fig:nav_t_vs_depth}}
 \caption{The effect of using different planning time, $T$, in DESPOT, compared with using $T = 0.25s$ in HyP-DESPOT, for navigation in a partially-known map.}\label{fig:nav_t_vs_all}

\end{figure}


\subsection{Effect of the Number of Scenarios}

Generally, problems with large $|\stateset|$'s can benefit significantly from the efficiency of HyP-DESPOT.
To perform robust planning, large $|\stateset|$-problems require more scenarios to effectively cover the states space and representative outcomes of actions and observations.
This creates many independent Monte Carlo simulations, thus increases parallelism in HyP-DESPOT. 
To study this effect, we analyze the performance of HyP-DESPOT on the navigation task (\secref{Section::Nav}), with $|\stateset|=169\times 2^{124}$, when using different \nscen's (100$\sim$5000) and 1 second planning time. \figref{fig:nav_k_vs_all} shows that the performance of HyP-DESPOT is highly scalable to large \nscen's. While the performance of DESPOT decays with \nscen, HyP-DESPOT achieves higher speed-up when sampling more scenarios, and thus generates higher quality solutions by searching larger trees. The search depth, however, decreases with \nscen, indicating that HyP-DESPOT searches a wider tree to produce robust decisions.

\subsection{Effect of the Planning Time}
Moreover, we experiment on the same navigation task by fixing \nscen and varying the planning time per step, $T$, and show that DESPOT takes much more time ($>40$x) to reach a comparable performance with HyP-DESPOT.
Particularly, we run HyP-DESPOT using $T=0.25s$, and set $T=1\sim10s$ for DESPOT. 
We use both the total discounted reward and the success rate of the robot reaching the goal within 60 steps to measure the performance.  
\figref{fig:nav_t_vs_all} shows that HyP-DESPOT significantly outperforms DESPOT in solution quality, by searching a larger and deeper tree, even when the latter uses 10s planning time. The performance gap decreases when DESPOT uses more time, but the trend becomes slow after $T=10s$.

\subsection{Effect of the Size of the Action Space}

\begin{figure}[!t]
\hspace*{0cm}  
\centering
\subfloat{\includegraphics[height=1.36in]{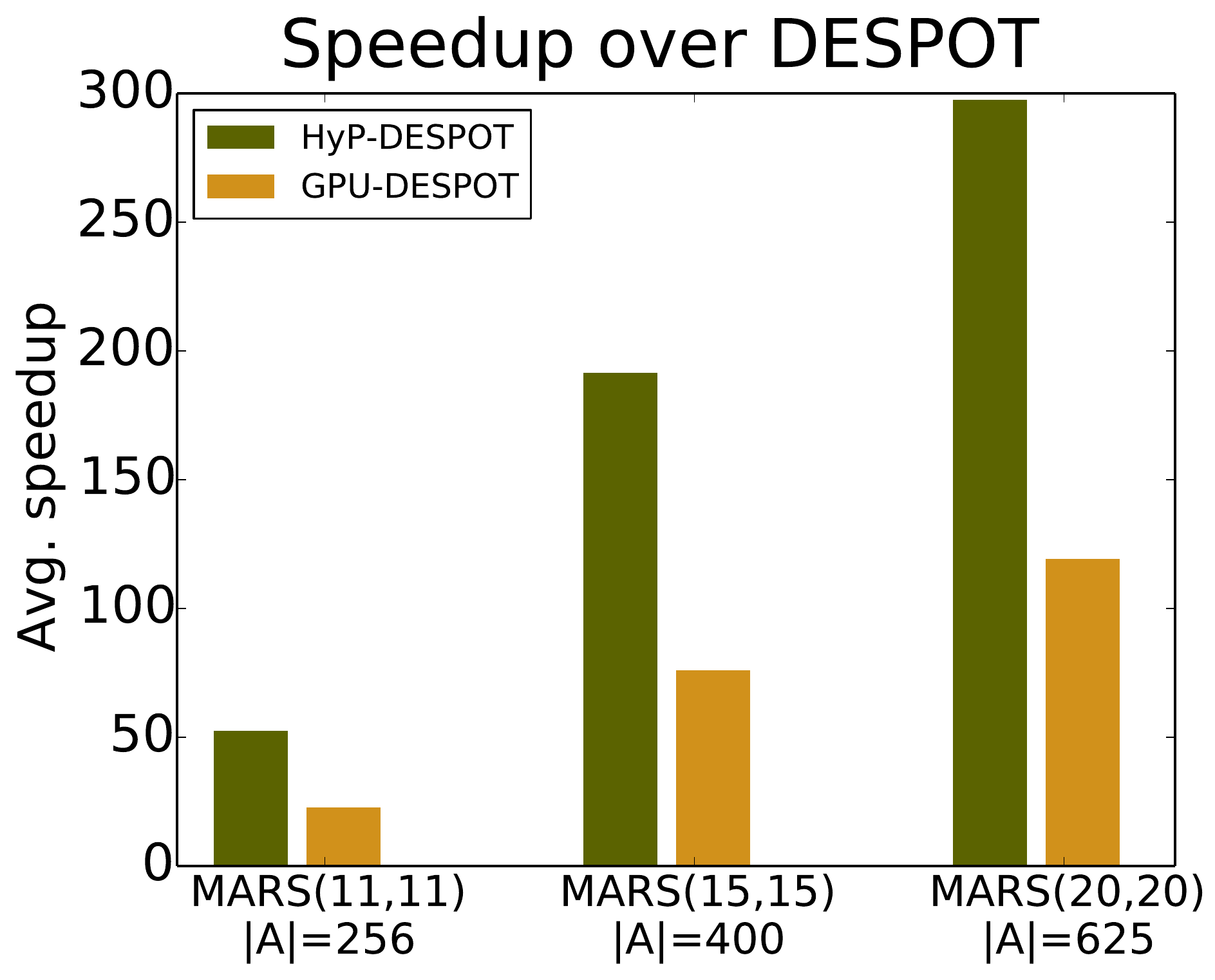}\label{fig:mars_probs_speedup}}
\hspace{0.01cm}
\subfloat{\includegraphics[height=1.36in]{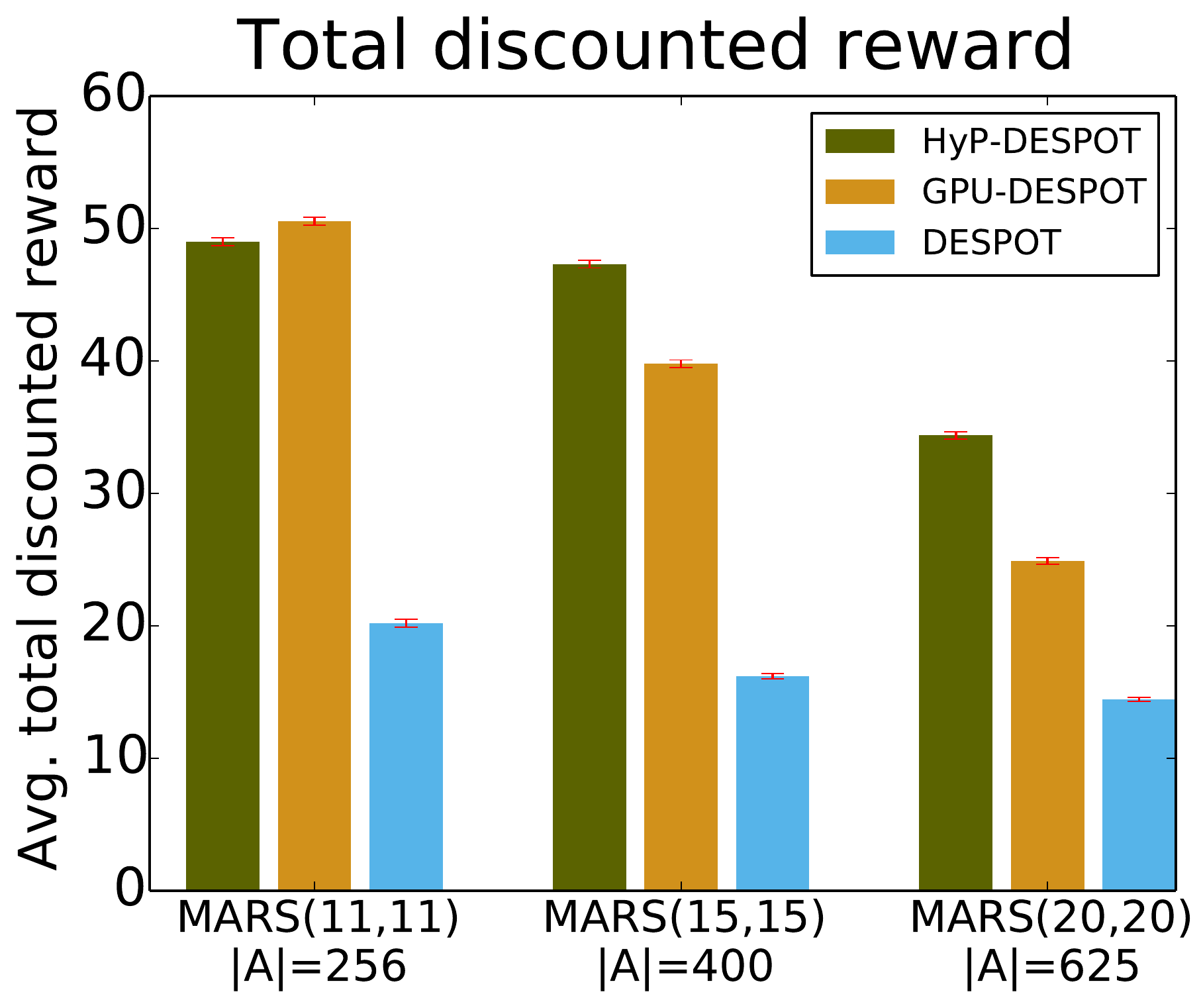}\label{fig:mars_probs_rw}} 
 \caption{The performances of HyP-DESPOT, GPU-DESPOT, and DESPOT in solving multi-agent \RockSample with different $|\actset|$ (256, 400, and 625).
     }\label{fig:marse_a_vs_all}
\end{figure}

Large $|\actset|$ improves the parallelism in Monte Carlo simulations like large \nscen's do. We illustrate that problems with large $|\actset|$'s can benefit more from HyP-DESPOT, by testing multi-agent \RockSample (11,11), (15,15), and (20,20), with $|\actset|$ to be 256, 400, and 625, respectively. We use a fixed $\nscen$ for all the tests. Results in \figref{fig:marse_a_vs_all} illustrate that HyP-DESPOT achieves higher speedup when $|\actset|$ increases, and generates significantly higher quality solutions than DESPOT in all the three tasks.

\subsection{Effect of the Number of Elements in Step Function}

\begin{figure}[!t]
\hspace*{0cm}  
\centering
\subfloat{\includegraphics[height=1.29in]{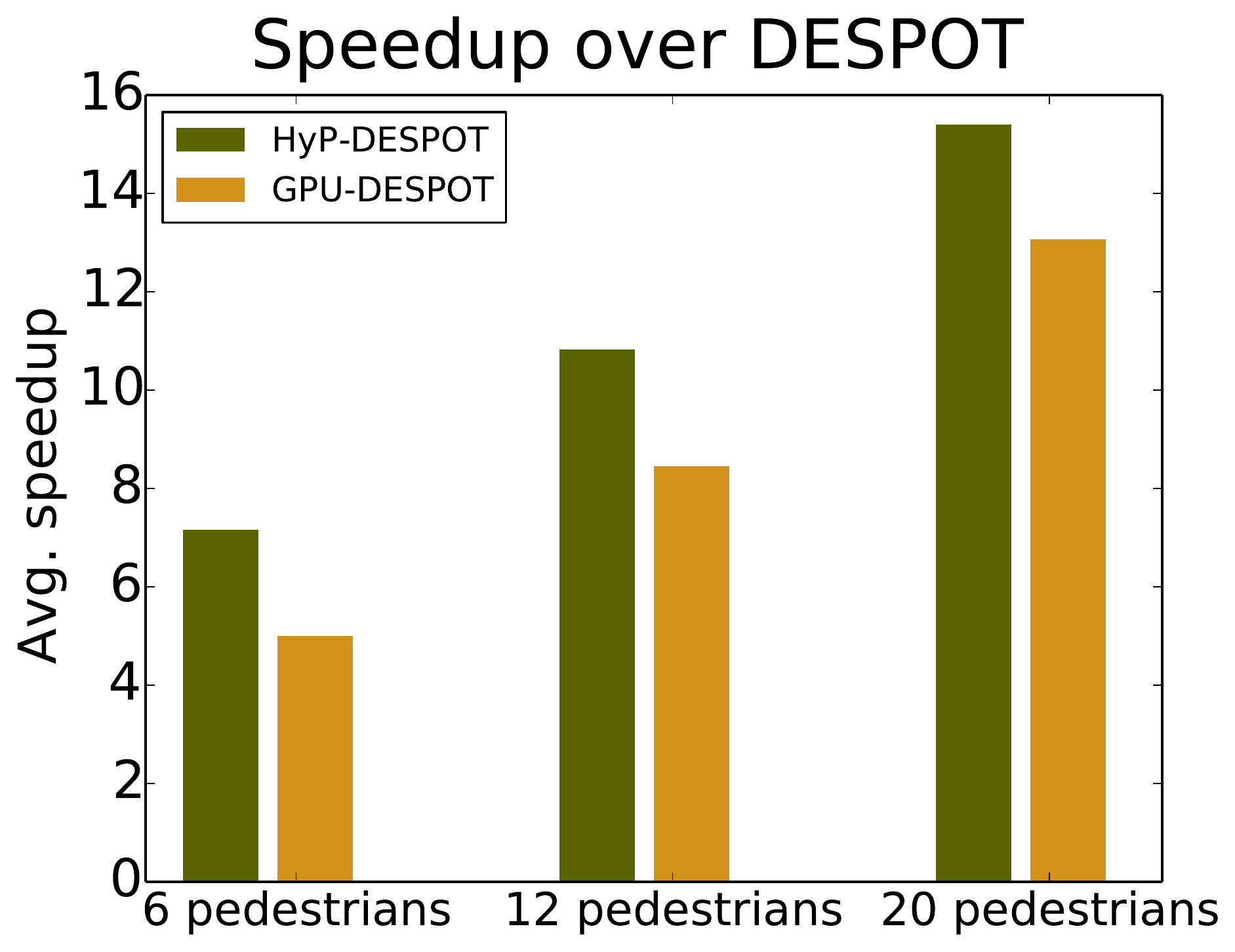}\label{fig:car_probs_speedup}}
\hspace{0.01cm}
\subfloat{\includegraphics[height=1.29in]{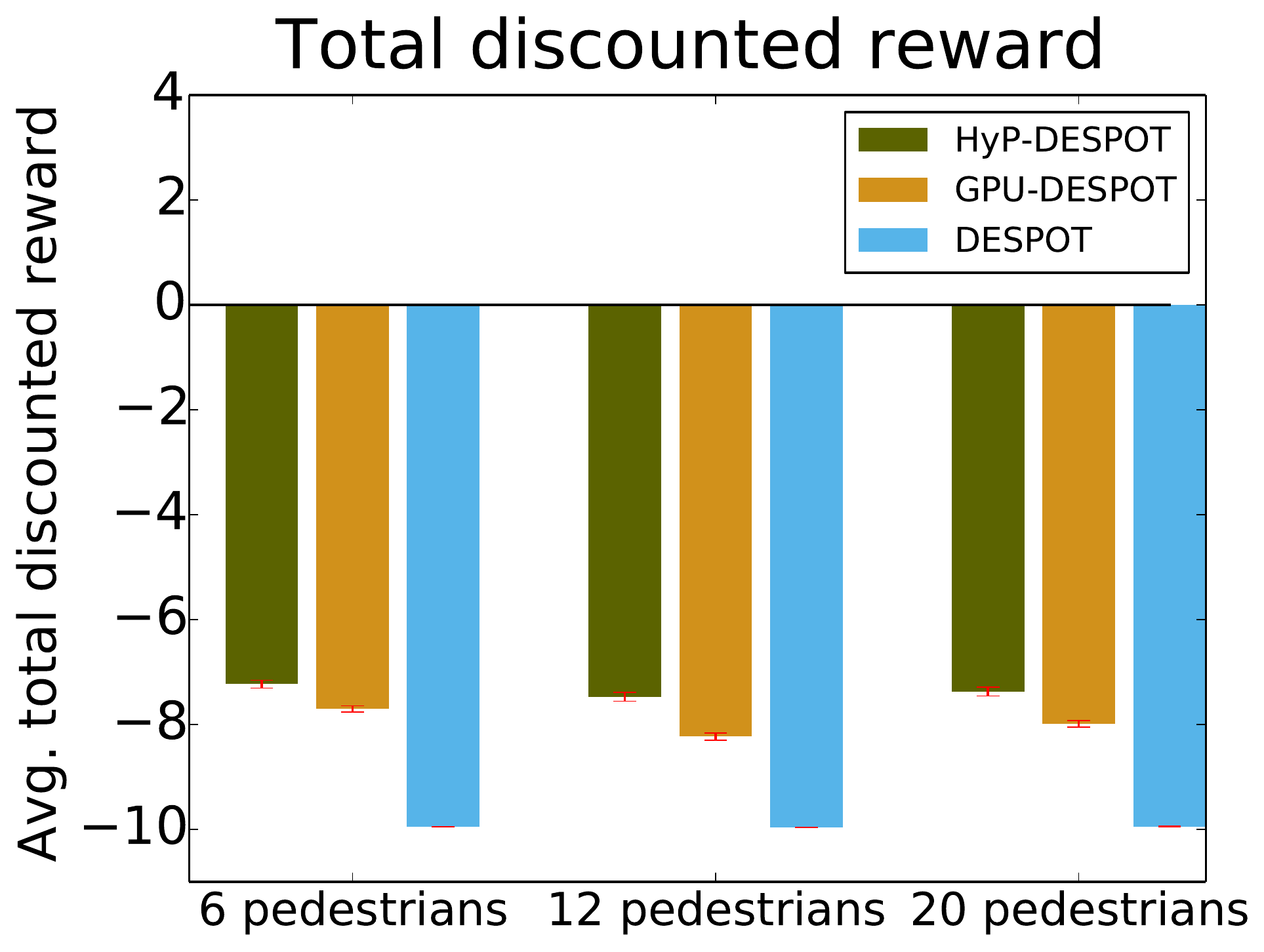}\label{fig:car_probs_rw}} 
 \caption{The performances of HyP-DESPOT, GPU-DESPOT, and DESPOT in solving the autonomous driving task, by considering different number of pedestrians (6, 12, and 20) in planning.
 }\label{fig:car_ped_vs_all}
\end{figure}

Large observation spaces, \obsset, however, usually restrict the performance gain from HyP-DESPOT by diverging scenarios into observation branches. Fortunately, many large-\obsset problems, \eg, driving among pedestrians (\secref{Section::CarDrving}), can still leverage the fine-grained \elementlevel parallelism to improve the performance. We illustrate that HyP-DESPOT becomes more efficient when a problem have more independent elements within a simulation step, by testing the effect of the number of pedestrians in the driving task. \figref{fig:car_ped_vs_all} shows that HyP-DESPOT achieves higher speedups when considering more pedestrians in planning, and brings significant improvements on the solution quality. 


\subsection{Experiments on an Autonomous Vehicle}\label{sec:scooter}
\begin{figure}[t]
\hspace*{0cm}  
\centering
\begin{tabular}{c@{}c}
\hspace*{-10pt}
\includegraphics[height=1.6in]{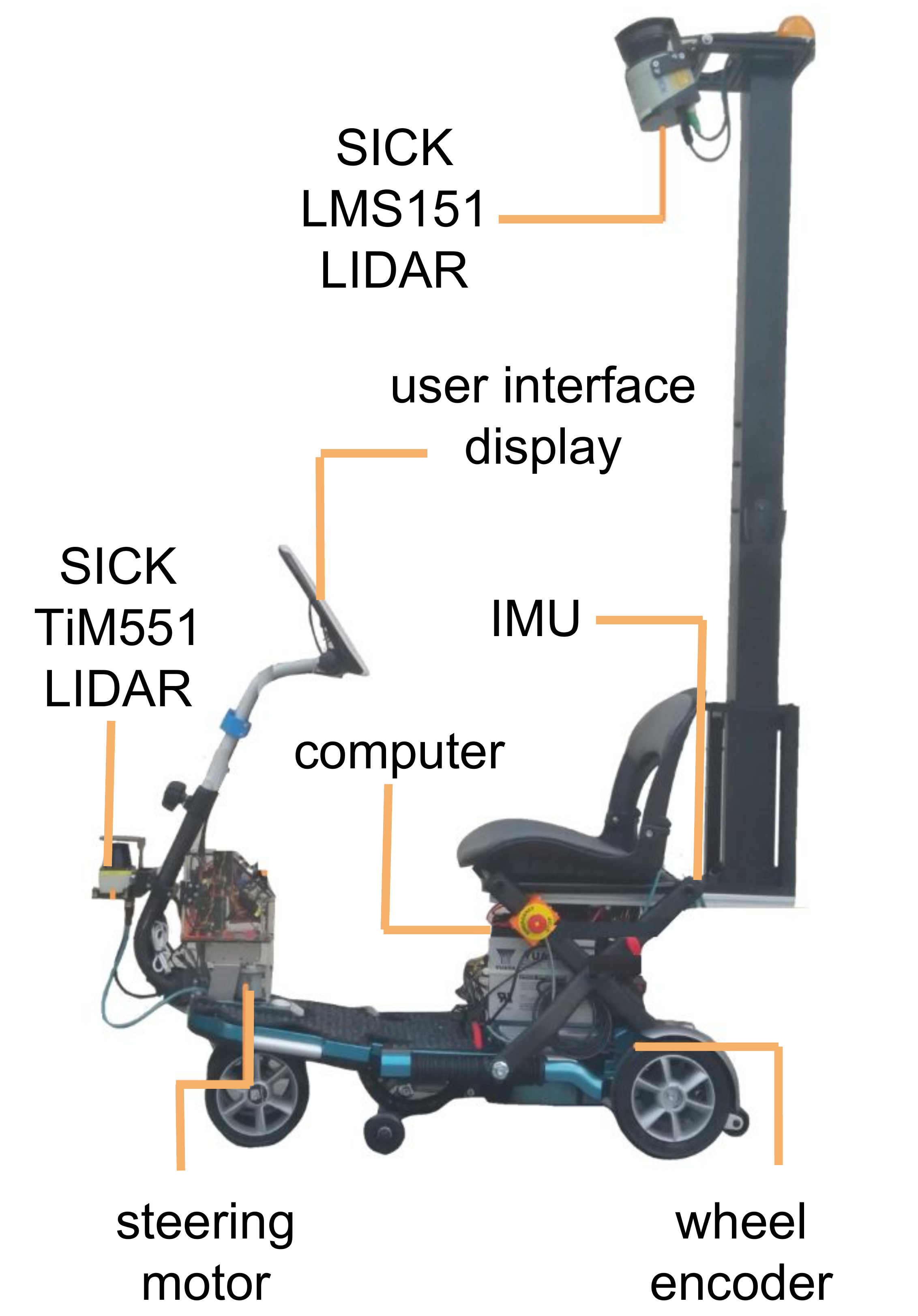}&
\includegraphics[height=1.6in]{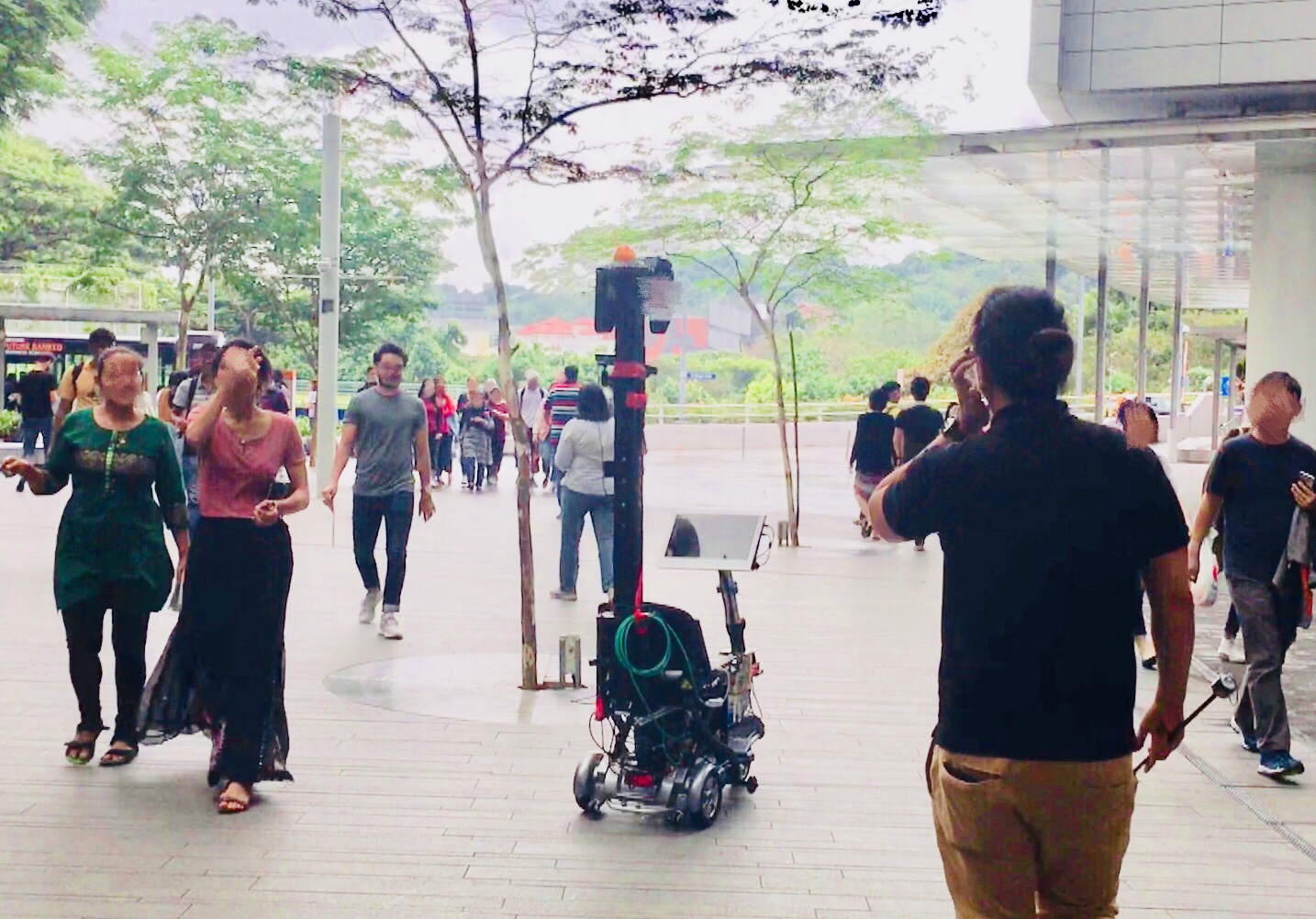} 
\end{tabular}
 \caption{The robot vehicle drives among pedestrians on a
   campus plaza. See also the accompanying video at \href{https://www.dropbox.com/s/39xi7k9hdhuyu9z/HyP-DESPOT-driving.mp4?dl=0}{https://www.dropbox.com/s/39xi7k9hdhuyu9z/HyP-DESPOT-driving.mp4?dl=0}.
     }\label{fig:robot-vehicle}
   \end{figure}

We implemented HyP-DESPOT on a robot vehicle for autonomous driving among
pedestrians on a campus plaza
(\figref{fig:robot-vehicle}).
The main vehicle  sensors consist of two LIDARs, an inertia measurement
unit (IMU), and wheel encoders. We use the SICK LMS151 LIDAR, mounted on top
of the vehicle, for pedestrian detection, and the SICK TiM551 LIDAR, mounted at
the front, for localization. The maximum vehicle speed is 1\,m/s.    HyP-DESPOT runs on an Ethernet-connected computer with an Intel Core
i7-4770R CPU running at 3.90 GHz, a GeForce GTX 1050M GPU (4GB VRAM), and 16 GB main memory.

We apply a two-level approach to control the vehicle
  \cite{Bai_2015}. At the
high level, we use the Hybrid A* algorithm \cite{stanley2006robot} to plan a
path. At the low level, we run HyP-DESPOT to compute the vehicle speed along
the planned path using the POMDP model described in
\secref{Section::CarDrving}.
The maximum planning time for HyP-DESPOT is 0.3s. So it replans
at approximately 3\,Hz.  

Our experiments on a campus plaza show that the autonomous vehicle can drive safely, efficiently, and smoothly, among many pedestrians. See the accompanying video \href{https://www.dropbox.com/s/39xi7k9hdhuyu9z/HyP-DESPOT-driving.mp4?dl=0}{https://www.dropbox.com/s/39xi7k9hdhuyu9z/HyP-DESPOT-driving.mp4?dl=0}
for an example.
\section{Conclusions}

This paper presents HyP-DESPOT, a massively parallel algorithm for online
planning under uncertainty. HyP-DESPOT performs parallel DESPOT tree search in
multi-core CPUs and massively parallel Monte Carlo simulations in GPUs.
When possible, it factors a complex system model and extracts fine-grained
parallelism for further performance gain.  By integrating CPU and GPU
parallelism in a multi-level scheme, HyP-DESPOT achieves
significant speedup over DESPOT on several large-scale planning tasks under
uncertainty.  The parallelization ideas underlying Hyp-DESPOT can be
generalized to other belief tree search algorithms, for instance, POMCP.

 HyP-DESPOT has two main performance bottlenecks for: communication
overhead among CPU threads and unbalanced workload of Monte Carlo simulations
in GPU threads.  Our next steps include implementing lock-free trees
to minimize communication among CPU threads and designing load balancing schemes
for GPU parallelization.








\newpage
\bibliographystyle{abbrvnat}
\bibliography{HyP-DESPOT}

\end{document}